\UseRawInputEncoding
\documentclass[11pt]{article}

\usepackage[preprint]{acl}
\usepackage{times}
\usepackage{latexsym}
\usepackage{amsmath}
\usepackage{adjustbox}
\usepackage{times}
\usepackage{latexsym}
\usepackage{amsfonts}
\usepackage{booktabs}
\usepackage[T1]{fontenc}

\usepackage[utf8]{inputenc}

\usepackage{microtype}

\usepackage{inconsolata}

\usepackage{graphicx}

\title{ChartREG++: Benchmarking and Improving Chart Referring Expression Grounding  under Diverse referring clues and Multi-Target Referring}

\author{
Tianhao Niu
\quad
Ziyu Han
\quad 
Xuan Dong
\quad 
Qingfu Zhu
\quad
Wanxiang Che \\
 Research Center for Social Computing and Interactive Robotics \\
 Harbin Institute of Technology, China \\
\texttt{\{thniu,car\}@ir.hit.edu.cn}
}

\usepackage{hyperref}
\usepackage{cleveref}
\usepackage[most]{tcolorbox}
\tcbuselibrary{listings,breakable}

\crefname{promptbox}{Box}{Boxes}
\Crefname{promptbox}{Box}{Boxes}

\NewTCBListing[
  auto counter,
  number within=subsection
]{promptbox}{!O{}}{
  enhanced,
  breakable,
  listing only,
  width=\linewidth,
  colback=black!2,
  colframe=black!20,
  boxrule=0.4pt,
  arc=1mm,
  left=2pt,
  right=2pt,
  top=4pt,
  bottom=4pt,
  boxsep=0pt,
  fonttitle=\bfseries,
  title={Box~\thetcbcounter: Full Prompt for Referring-Expression Generation},
  label type=promptbox,
  #1,
  listing options={
    basicstyle=\rmfamily\scriptsize,
    breaklines=true,
    breakatwhitespace=false,
    columns=fullflexible,
    keepspaces=true,
    showstringspaces=false
  }
}

\usepackage{booktabs}
\usepackage{array}
\usepackage{tabularx}
\begin{document}
\maketitle
\begin{abstract}
    Referring expression grounding is a core problem in visual grounding and is widely used as a diagnostic of spatial grounding and reasoning in vision–language models, yet most prior work focuses on natural images. In contrast, existing chart referring expression grounding-related benchmarks remain limited: (1) they largely adopt bounding boxes, constraining localization precision for fine chart elements; (2) they mostly assume one and two referred target instances, failing to handle multi-instance target references; (3) the language expressions over-rely on textual cues or data-ranking clues in the chart; (4) they cover only a narrow range of chart types.
    To address these issues, we introduce a chart referring expression grounding benchmark that first systematically supports multiple localization formats, multiple referred targets, diverse grounding cues  and diverse chart types. Results across representative multi-modal large models reveal a significant performance gap. 
    We further introduce an LLM-free synthesis pipeline that exploits the inherent alignment between matplotlib internal artist hierarchy and rendered chart primitives to derive pixel-accurate, multi-granularity instance segmentation masks. We train an instance segmentation model with the synthesized masks and integrate it into a general-purpose multi-modal grounding framework. The resulting system consistently outperforms baselines on our benchmark and generalizes well to a ChartQA-derived real-chart grounding benchmark.
\end{abstract}

\section{Introduction}
\begin{table}[t]
\centering
\scriptsize
\setlength{\tabcolsep}{3.0pt}
\renewcommand{\arraystretch}{1.05}
\resizebox{\columnwidth}{!}{
\begin{tabular}{@{}lccccc@{}}
\toprule
Benchmark & Task & Loc. & Clues & \#Elem. & Avg. tgt. \\
\midrule
InfoDet & Elem. Loc. & BBox & -- & 26 & -- \\
RefChartQA & Attr. QA & BBox & T/D$^\dagger$ & 3 & 2.1 \\
ChartLens & Attr. QA & BBox & T/D$^\dagger$ & 3 & 1.8 \\
ChartRef & Ref. Ground. & BBox & T & 11 & 1.0 \\
DOGR (Chart) & Attr. QA & BBox & -- & 1 & -- \\
\textsc{ChartREG++} & Ref. Ground. & P/B/S & T/D/V & \textbf{18} & \textbf{9.7} \\
\bottomrule
\end{tabular}
}
\caption{
Comparison with chart visual grounding-related benchmarks.
P/B/S denotes point, bounding box, and segmentation mask.
T/D/V denotes textual, data, and visual referring clues.
$^\dagger$ indicates that data clues are dominated by data-rank cues.
}
\label{tab:intro_benchmark_comparison}
\end{table}
\label{sec:intro}
With the rapid development of multimodal foundation models \cite{liu2023visual,bai2025qwen3vltechnicalreport,wang2025internvl35advancingopensourcemultimodal,10655294}, chart understanding tasks—such as chart question answering \cite{wang2024charxiv,masry-etal-2022-chartqa,tang2025chartmuseum}, chart parsing \cite{li2026visualselfrefinepixelguidedparadigm}, and chart-to-text  generation\cite{kantharaj-etal-2022-chart} —have seen notable progress. Yet many early systems produce only a textual answer or description, without explicitly linking the answer to its supporting evidence in the chart. This limits direct verification by users. To build trustworthy chart understanding and generation systems, we need stronger interpretability and verifiability: a model should not only provide a conclusion, but also localize the visual evidence in the original chart that justifies the conclusion.

Driven by this goal, recent work has begun to incorporate fine-grained chart grounding signals into model outputs. \textbf{The first line of work} injects visual grounding information into multimodal reasoning traces to improve chart QA performance (e.g., PointRFT \cite{ni2025pointrft} and ChartPoint \cite{xu2025chartpointguidingmllmsgrounding}). However, these methods mainly target QA answer accuracy and do not explicitly evaluate the ability to reference fine-grained visual evidence for chart elements. \textbf{The second line of works are attribute QA} (e.g., RefChartQA \cite{vogel2025refchartqagroundingvisualanswer}, ChartLens \cite{suri-etal-2025-chartlens}, and DOGR \cite{zhou2025dogrversatilevisualdocument}) which requires a model to output both the answer and the corresponding evidence regions in the chart \textbf{The third line of works are referring expression grounding.} ChartRef \cite{tjandrasuwita2025chartref}, introduces a benchmark for chart-domain referring expression grounding, where the model directly maps a natural-language expression to the corresponding chart region without producing any textual answer. We name the second and third line works as referring-expression-grounding-related tasks.\footnote{We observe that, in the second-line setting, the “direct evidence” is often already specified or strongly constrained by the question text itself. Compared with the third-line setting, the difference is mainly the surface form of the expression and the additional requirement to output an answer. We therefore group both settings as referring-expression-grounding-related tasks.}

Despite this progress, existing chart referring-expression-grounding-related benchmarks exhibit at least one of the following limitations: (1) evidence is commonly represented by bounding boxes, which is poorly suited for irregular or thin chart elements such as error bars and lines; (2) most queries refer to only one or two targets, leaving multi-target expressions underexplored; (3) most queries overemphasize textual/localization cues or data-ranking based clues to specify the target, covering a narrow set of referring cues; (4) the coverage of chart element types is still limited.

To address these issues, we propose ChartREG++, a chart referring expression grounding benchmark that first jointly supports diverse referring cues, multiple referred targets, a broader range of element types, and multiple localization formats. 
Firstly, the referring expressions in ChartREG++ are driven by \textbf{\emph{three categories of referring cues, covering 13 subtypes.}} We also include expressions that combine multiple cue types.
Secondly, to improve the element type coverage while ensuring similarity to real-world as possible, we create ChartREG++ based on two datasets that are close to real charts ECDBench \cite{yang2025effectivetrainingdatasynthesis} and ChartMimic \cite{yang2025chartmimic}. The result benchmark contains \emph{\textbf{18 chart element types}} across different granularities and on \emph{\textbf{average 9.7 referring targets}}, which \emph{\textbf{jointly supports point-based, bbox-based, and mask-based}} Chart referring grounding. \footnote{We summarize more detailed statistics of ChartREG++ and its differences from prior benchmarks in Table~\ref{tab:benchmark_comparison_2} and Table~\ref{tab:benchmark_comparison_1}.}Experiments with strong multimodal foundation models and visual grounding models indicate substantial room for improvement on ChartREG++.



We further introduce an LLM-free mask synthesis pipeline for chart instance segmentation. \emph{\textbf{On the one hand}}, we exploit the alignment between Matplotlib’s Artist hierarchy and rendered primitives to generate multi-type masks at multiple granularities, from low-level primitives to higher-level chart components. \emph{\textbf{On the other hand}}, we use the Artist semantics and API context to disambiguation of visually similar elements such as line markers and independent scatter points and assign fine-grained mask labels for each instance. \emph{\textbf{Finally}}, the method is scalable as it operates directly on executable Matplotlib code and is LLM-free.
We discuss the relationship to existing synthesis methods in Sec~\ref{sec:method}. Using the synthetic data, we train an instance segmentation model to generate mask candidates and combine them with the visual grounding framework using  Set-of-Mark prompting. The framework achieves significant gains on ChartREG++ compared with the baselines, while still leaving clear headroom. The framework also achieves the better performance in our experiments on multiple dimensions of the ChartQA real-chart subset within the ChartLens benchmark which contains non-matplotlib chart images.

We summarize our contributions as follows \footnote{We discuss more about the related works in Appendix~\ref{sec:related_work}} \footnote{More benchmark examples are shown in Appendix~\ref{subsec:benchmark_examples}}:
\begin{itemize}
\item{We introduce ChartREG++, a chart referring expression grounding benchmark that first jointly supports diverse referring cues, multiple referred targets, diverse chart element types, and grounding in point / bounding box / segmentation formats.}
\item{We evaluate strong multimodal foundation models and visual grounding models, showing that ChartREG++ remains challenging and offers substantial room for improvement.}
\item We further introduce an LLM-free mask synthesis pipeline for chart instance segmentation. Models trained with our synthetic supervision deliver strong improvements on ChartREG++ and competitive results on the non-matplotlib ChartLens-ChartQA.
\end{itemize}

\begin{figure*}[t]
  \centering
  \includegraphics[width=\textwidth]{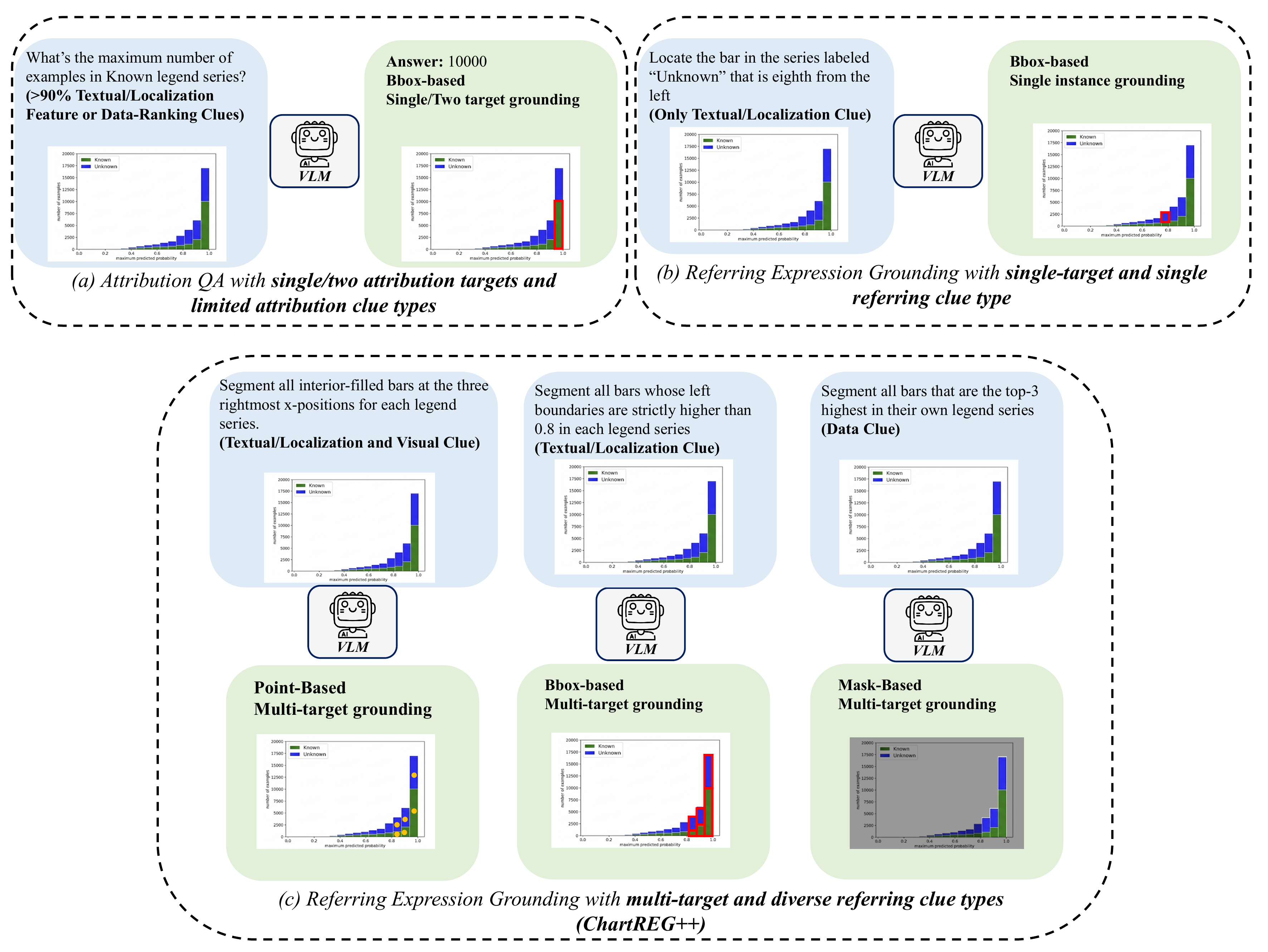}
  \caption{Comparison between ChartREG++ (c) and prior benchmarks. Prior work (a), such as RefChartQA\cite{vogel2025refchartqagroundingvisualanswer} and ChartLens\cite{suri-etal-2025-chartlens}, evaluates attribution-aware chart question answering, while (b) ChartRef\cite{tjandrasuwita2025chartref} evaluates the ability to link natural language to chart image elements. In these benchmarks, referred targets are mostly identified from textual/location cues in the expression or simple ranking cues in the data, and the target set usually contains only one or two instances. They also mainly rely on bounding boxes for localization.}
  \label{fig:benchmark_compare}
\end{figure*}

\section{Task Formulation}

\label{sec:task_formulation}
Given a chart image $I \in \mathbb{R}^{H \times W \times 3}$ and a referring expression $q$, 
the goal is to localize all chart elements referred to by $q$.
\textbf{Here we assume that the target elements are of the same type.} 
Note that the \emph{homogeneous-type} assumption does not restrict targets to the same low-level drawing primitive or path representation. Instead, we define an \emph{element type} as a semantically coherent unit aligned with human chart perception, and we explicitly expose a multi-granularity hierarchy within each type (e.g., a boxplot can be referred to at the level of \textit{Full Box} which we refer to as box along with its whispers, caps and median line in our benchmark as a holistic semantic object, or at finer levels such as whiskers, median line, and caps). As a result, the homogeneous-type constraint still covers the common \emph{whole$\leftrightarrow$part} references used by humans; meanwhile, more complex heterogeneous composite references (e.g., ``the tallest bar and the line marker on it'') can often be naturally decomposed into a sequence of verifiable sub-references (first localize the primary object, then localize its related components or relational counterparts).

Formally, each sample is $(I,q,c,\mathcal{Y})$, where $I \in \mathbb{R}^{H \times W \times 3}$ is a chart image, $q$ is a referring expression, and $c \in \mathcal{C}$ denotes the \emph{single} chart element type referred to by $q$ (homogeneous-type targets).
The ground-truth annotation is an unordered set $\mathcal{Y}=\{g_i\}_{i=1}^{K}$, where $K$ is the number of referred targets and each $g_i$ specifies the localization of a target of type $c$.
ChartREG++ supports three localization formats: (i) a point $g_i^{\text{pt}}=(x_i,y_i)$, (ii) a bounding box $g_i^{\text{box}}=(x_i^{\min},y_i^{\min},x_i^{\max},y_i^{\max})$, or (iii) an instance mask $g_i^{\text{seg}} \in \{0,1\}^{H \times W}$.
A model $f_{\theta}$ predicts a set of localizations $\hat{\mathcal{Y}} = f_{\theta}(I,q)$ in the same format.

\section{ChartREG++ Benchmark}
\label{sec:benchmark_construction}

\subsection{Benchmark Construction}
This section introduce our benchmark construction pipeline. More details are in Appendix~\ref{sec:benchmark_details}. Benchmark examples are in Appendix~\ref{subsec:benchmark_examples}.
\subsubsection{Creating taxonomy}
\paragraph{Creating Referring target element type taxonomy} 
We define 18 common chart element types in Figure~\ref{fig:ds_distribution_all} bottom right. These elements cover different referring granularities; for example, for Box we annotate not only the complete box (including its caps, box patch, and median line), but also more atomic parts such as the box patch and the median line. Examples are shown in Appendix~\ref{subsec:benchmark_examples}.

\paragraph{Creating Referring clue type taxonomy}
We group referring cues into three primary categories: \textbf{Data Feature} (Value-Range Filtering; Rank-Band Set Selection; Local-Structure Patterns; Cross-Series Relations), \textbf{Visual Feature} (Color Attributes; Shape Style; Line Style / Stroke Style; Fill Style), and \textbf{Textual Localization  Feature} (Axis labels; Non-data axis tick values and positions; Legend entries and their positions; Subplot titles, identifiers and positions; Text annotations directly on the chart). For referring expressions driven by each primary cue, we ensure that confirming the referred targets must rely on that primary cue.    In addition, to better mimic real user queries, we include hybrid feature referring expressions that combine two primary cue categories; for these \textbf{hybrid features}, we do not enforce that each involved primary cue category individually contributes to uniquely determining the targets. Examples are shown in Appendix~\ref{subsec:benchmark_examples}.\footnote{Nevertheless, more than 50\% of the samples of such feature still require jointly leveraging the involved primary cues for localization.}

\subsubsection{Collecting Chart Images}
Our chart images are sampled from ChartMimic and ECDBench in equal proportions. ChartMimic consists of human-curated (figure, instruction, code) triplets collected from authentic chart use cases in scientific papers. ECDBench is built via a carefully designed synthesis-and-review pipeline that filters and fixes rendering issues by editing and re-rendering plotting code, and it is used to evaluate transfer to real-world chart benchmarks. Together, these two sources provide plotting code and reflect the diversity and complexity of real charts.

\begin{figure*}[t]
  \centering
  \includegraphics[width=\textwidth]{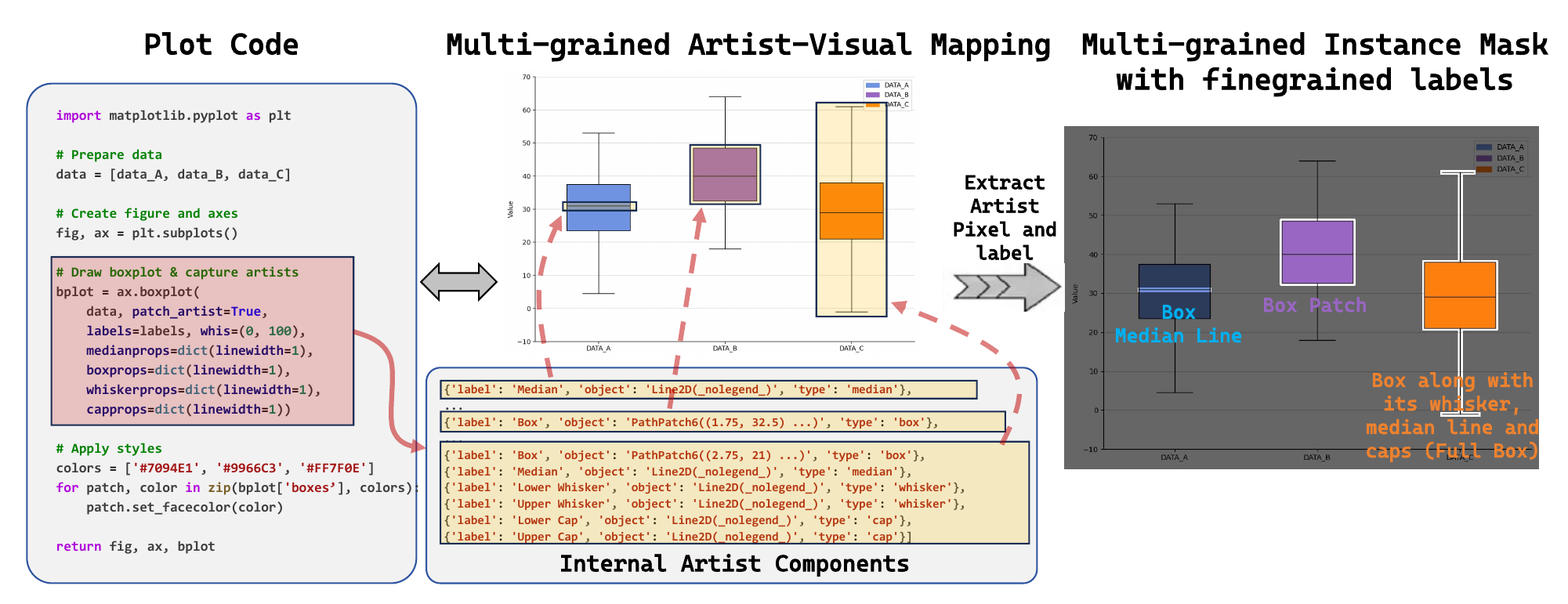}
  \caption{Proposed pipeline for multi-granularity instance masks with fine-grained chart-element labels. We start from large-scale Matplotlib plotting code collected from the web or synthesized at scale, and trace each plotting API call to the rendered \emph{Artist} objects together with their associated metadata. Using the Artist hierarchy, we construct a multi-granularity \emph{Artist-to-visual} mapping that links code-level primitives to image regions. We then recover instance masks by extracting the pixels contributed by each Artist at different granularities (e.g., a box median line, a box patch, and a complete boxplot instance including the box, whiskers, caps, and median line). Finally, we assign fine-grained mask labels by combining Artist metadata with API semantics, 
therefore each mask is described by an specific semantic role (e.g., the \emph{median line} \emph{of a boxplot instance}) instead of a coarse element type (e.g. line)}
  \label{fig:syntheisze_pipeline}
\end{figure*}
\subsubsection{Generating Referring Expressions}
For each referred target element type, we do the following. We randomly sample images that contain this element type, and ensure that the images selected for later types do not overlap with those already used. We then provide the corresponding plotting code to a large model to generate referring expressions. For each primary clue category or hybrid feature setting, we prompt the model to produce 5-10 different candidate referring expressions that \emph{use different sub-clue types as much as possible, and then randomly select one expression from the candidates for each primarily clue and hybrid feature category}.\footnote{In our preliminary trials, large models often fail to generate accurate referring expressions. We therefore enforce the subject name in the prompt (e.g., for a complete marker line series, the subject must be “line series along with its markers”), and require each expression to start with “the/all [subject]”. We acknowledge the potential risk of template-like phrasing, and leave more diverse expressions as future work. However, since \textbf{we use multiple LLM to generate expressions and forcing the LLM to generate descriptions that contain different sub-cues for the 5-10 candidates , which reduce bias of single LLM and increase diversity.}} To reduce potential bias from any single model, we split the images to be annotated into three equal subsets, and apply the same procedure using Gemini3-Pro-Preview \cite{comanici2025gemini25pushingfrontier}, GPT-5.2 \cite{openai2024gpt4technicalreport}, and DeepSeekV3.2 \cite{deepseekai2025deepseekv32pushingfrontieropen}, respectively.

\subsubsection{Generating Referring targets}
We use DeepSeekV3.2 to synthesize the referred target elements conditioned on the plotting code and the referring expression. Specifically, for the referred element type associated with the expression, we first identify the corresponding API lines in the code, and then prompt the model to select the referred targets from these lines. We then generate instance masks for the corresponding referred element category using the method in \ref{sec:method}, conditioned on the synthesized referred-target information, and then obtain Bounding Box annotations by converting each mask to its tight enclosing box

\subsubsection{Manual filtering and verification.}

We manually check, refine and verify each sample with three criteria.
\textbf{(i) Referring expression precision:} the expression refers to the intended target set, and it must rely on the specified \emph{primary clue category}.\footnote{We ensure that, after removing the specified primary clue category, the expression can no longer refer to the original target set.}
\textbf{(ii) Uniqueness (non-ambiguity):} the expression does not refer to any other objects beyond the intended targets.
\textbf{(iii) Target correctness:} the selected targets indeed correspond to what the expression describes.
The quality control procedure consists of two rounds:
In the first round, two undergraduate annotators with a computer-science background independently review every sample, flag violations of the above criteria, and provide concrete revision suggestions (i.e.., rewriting the expression or correcting the target set).
In the second round, a Ph.D.-level annotator re-checks these suggestions and applies the finalized revisions.
After the updates, the Ph.D annotator and one of the first-round annotators conduct an additional full pass over the dataset to ensure that every sample satisfies the three criteria.

\subsection{Benchmark Statistics and Distribution}


As shown in Table~\ref{tab:benchmark_comparison_2} and Table~\ref{tab:benchmark_comparison_1}, \textsc{ChartREG++} contains 18 referred element types across different granularities, with 850 chart images, 3.4K referring expressions, and an average of 9.7 referred targets per expression. Note that we prioritize coverage of grounding phenomena over repeated enumeration of similar instances. We discuss more about this in Appendix~\ref{subsec:benchmark_scale_discussion}.

Beyond these basic statistics, \textsc{ChartREG++} also exhibits diverse data distributions. 
First, it covers charts with a wide range of visual and structural complexity, where the underlying plotting scripts contain 20--280 lines of code, and includes both single-subplot and multi-subplot cases (Figure~\ref{fig:ds_distribution_all}, top left). 
Second, it has a relatively balanced distribution of referring-expression lengths (Figure~\ref{fig:ds_distribution_all}, top middle). 
Third, it includes many multi-target expressions: expressions referring to one or two targets account for only 50\% of the dataset, compared with 80\% in \textsc{ChartLens} and 90\% in \textsc{RefChartQA} (Figure~\ref{fig:ds_distribution_all}, top right). 
Finally, \textsc{ChartREG++} maintains broad coverage over both cue categories and referred element types, as shown in Figure~\ref{fig:ds_distribution_all} bottom left and bottom right, respectively.

\section{Method}
\label{sec:method}

We adopt a \emph{detect-select} two-stage pipeline to improve performance on our benchmark. The first stage generates candidate masks using an instance segmentation model, and the second stage applies \textsc{Set-of-Mark} prompting to let an MLLM directly select the matched mask set from the candidates. The key challenge is how to obtain large-scale, stable instance-level mask supervision that covers different element granularities at a reliable cost.

To address this challenge, we propose a synthesis pipeline that automatically generates multi-type and multi-granularity chart instance masks from executable matplotlib plotting scripts. 
\emph{\textbf{Firstly}}, the method exploits Matplotlib's internal Artist hierarchy to trace rendered chart elements in a structured manner. 
This allows us to synthesize masks along a primitive--part--composition hierarchy, covering both basic visual primitives and higher-level chart components. 
\emph{\textbf{Secondly}}, the method combines Artist semantics with API context to produce fine-grained and disambiguated mask labels. 
For instance, markers with similar visual appearances can be assigned different labels depending on whether they are created by \texttt{ax.plot} as part of a line series or by \texttt{ax.scatter} as standalone scatter points. 
Such semantic tracing provides more reliable supervision than appearance-only mask extraction. 
\emph{\textbf{Finally}}, since the pipeline requires neither manual annotation nor LLM-based labeling and only depends on executable Matplotlib code, it can naturally scale to large collections of web-sourced or automatically synthesized plotting scripts. 
An example is shown in Figure~\ref{fig:syntheisze_pipeline}.

Compared with existing automated chart annotation works, our method targets \emph{instance masks} with \emph{multiple granularities} and \emph{fine-grained labels}, while prior pipelines mainly focus on detection-style bounding boxes and are less suited for stable, fine-grained instance mask supervision for chart primitives and parts.
InfoDet synthesizes infographics and uses an integrated SVG parser to extract bounding boxes from the rendered SVG. However, SVG typically exposes low-level drawing directives (e.g., \texttt{path}/\texttt{g}) with limited built-in semantics; these primitives are not naturally aligned with human semantic units (e.g., a complete box in a boxplot), which requires additional aggregation rules. Moreover, visually similar primitives may correspond to different semantics, which is difficult to disambiguate reliably from geometry alone, limiting the construction of a stable fine-grained label system.
CHARTREF leverages Matplotlib scripts but relies on a large language model to infer data semantics and then programmatically extract bounding boxes, keeping box-based supervision and introducing model dependency.
DOGR mainly targets document grounding/referring with \emph{text} bounding boxes, rather than instance segmentation supervision for chart primitives/parts.

\subsection{Point and Box Grounding}
\subsubsection{Models}
We evaluate the zero-shot grounding ability of representative multimodal foundation models on \textsc{ChartREG++}, covering both closed-source and open-source systems (e.g., Gemini3-Pro \cite{comanici2025gemini25pushingfrontier}, GPT-5.2 \cite{openai2024gpt4technicalreport}, Molmo-v2-8B \cite{clark2026molmo2openweightsdata}, the Qwen3-VL \cite{bai2025qwen3vltechnicalreport} family, and InternVL3.5-8B \cite{wang2025internvl35advancingopensourcemultimodal}).
For closed-source models, we require outputs in a normalized $[0,1000]$ coordinate system as \texttt{point} or \texttt{bbox}.
For open-source models, we design prompts to follow their released referring-localization style as closely as possible. We use greedy sampling for all models.

\subsubsection{Evaluation metrics}
\paragraph{Point-based grounding.}
A predicted point is counted as a match to a target instance if it falls inside the target instance mask; otherwise, we compute the Euclidean distance from the point to the nearest foreground pixel of the mask, and treat it as a match if the distance is $\leq 5$ pixels.\footnote{We choose the r as the maximal image resolution is 3200.}
Under a one-to-one constraint, we apply Hungarian matching to obtain the maximum matching, and compute per-sample Precision/Recall/F1.
We report macro-P/R/F1 averaged over all test samples.
\paragraph{BoundingBox-based grounding.}
We compute pairwise IoU between predicted boxes and ground-truth boxes, and consider a pair eligible if $\mathrm{IoU}\ge 0.5$.
Under a one-to-one constraint, we apply Hungarian matching to obtain the globally optimal assignment, and compute per-sample Precision/Recall/F1.
We report macro-P/R/F1 averaged over all test samples.


\begin{table}[t]
\centering

\footnotesize
\begin{adjustbox}{max width=1.0\columnwidth}
\begin{tabular}{@{}lcccccc@{}}
\toprule
& \multicolumn{3}{c}{\textbf{ECDBench}} 
& \multicolumn{3}{c}{\textbf{ChartMimic}} \\
\cmidrule(lr){2-4}\cmidrule(l){5-7}
\textbf{Model} 
& \textbf{P} & \textbf{R} & \textbf{F1} 
& \textbf{P} & \textbf{R} & \textbf{F1} \\
\midrule
\multicolumn{7}{@{}l}{\textbf{Point-Based}} \\
Gemini3-Pro    
& \textbf{32.9} & \textbf{39.4} & \textbf{33.4} 
& 36.8 & 44.8 & 37.0 \\
GPT-5.2        
& 18.3 & 24.2 & 18.7 
& 31.7 & 40.3 & 30.5 \\
Molmo-v2-7B    
& 31.0 & 33.6 & 30.0 
& \textbf{46.9} & \textbf{51.7} & \textbf{45.2} \\
Qwen3-VL-8B    
& 8.4 & 10.6 & 8.5 
& 10.5 & 13.4 & 10.3 \\
\midrule
\multicolumn{7}{@{}l}{\textbf{BBox-Based}} \\
Gemini3-Pro    
& \textbf{24.2} & \textbf{23.9} & \textbf{23.9} 
& \textbf{30.3} & \textbf{29.8} & \textbf{29.8} \\
GPT-5.2        
& 10.5 & 10.4 & 10.4 
& 23.3 & 23.0 & 23.0 \\
Qwen3-VL-7B    
& 8.6 & 7.9 & 8.1 
& 15.6 & 14.9 & 14.9 \\
Qwen3-VL-32B   
& 15.0 & 14.8 & 14.6 
& 15.8 & 15.7 & 15.5 \\
InternVL3.5-8B 
& 2.2 & 2.0 & 2.0 
& 5.8 & 5.4 & 5.5 \\
\bottomrule
\end{tabular}
\end{adjustbox}
\caption{Point- and box-based grounding performance (P/R/F1, \%) on ChartREG++}
\label{tab:bbox_point_main_res}
\end{table}
\subsubsection{Results}
As shown in Table~\ref{tab:bbox_point_main_res}, existing strong open-source and closed-source models still have substantial room for improvement on chart-domain referring expression grounding.
Even the best-performing model achieves F1 below 50\% on both point-based and bbox-based settings, highlighting the challenge of our benchmark.
We also observe that bbox-based performance is consistently lower than point-based performance, suggesting that producing \emph{verifiable geometric evidence} remains difficult in our task.

\begin{table*}[t]
\centering
\small
\begin{adjustbox}{max width=\textwidth}
\begin{tabular}{@{}lccccccc@{}}
\toprule
& \multicolumn{3}{c}{\textbf{ECDBench}} 
& \multicolumn{3}{c}{\textbf{ChartMimic}} 
& \textbf{mIoU$_{\text{union}}$} \\
\cmidrule(lr){2-4}\cmidrule(lr){5-7}\cmidrule(lr){8-8}
\textbf{Model} 
& \textbf{P} & \textbf{R} & \textbf{F1} 
& \textbf{P} & \textbf{R} & \textbf{F1} 
& \textbf{(\%)} \\
\midrule

\multicolumn{8}{@{}l}{\textit{Zero-shot E2E grounding}} \\
Sa2VA-Qwen2.5-VL-7B   
& -- & -- & -- & -- & -- & -- & 21.00 \\
Sa2VA-InternVL3-14B   
& -- & -- & -- & -- & -- & -- & \textbf{24.00} \\

\midrule
\multicolumn{8}{@{}l}{\textit{Zero-shot two-stage grounding with SAM3 candidates}} \\
SAM3-SoM-QwenVL-32B               
& \textbf{28.18} & \textbf{24.22} & \textbf{25.01} 
& \textbf{29.55} & \textbf{21.55} & \textbf{22.91} 
& \textbf{22.13} \\
SAM3-SoM-InternVL3.5-30B-A3B      
& 24.57 & 21.44 & 21.84 
& 24.05 & 18.39 & 19.16 
& 18.79 \\

\midrule
\multicolumn{8}{@{}l}{\textit{Synthetic-supervised two-stage grounding with our candidates}} \\
Ours-SoM-QwenVL-32B               
& \textbf{44.64} & \textbf{52.55} & \textbf{43.82} 
& \textbf{43.75} & \textbf{49.85} & \textbf{43.79} 
& \textbf{42.93} \\
Ours-SoM-InternVL3.5-30B-A3B      
& 34.60 & 35.58 & 34.25 
& 41.81 & 40.88 & 37.29 
& 34.70 \\

\midrule
\multicolumn{8}{@{}l}{\textit{Category-filtered and oracle analysis}} \\
Ours-SoM-InternVL3.5-8B           
& 43.54 & 35.76 & 37.07 
& 48.94 & 41.52 & 42.07 
& 36.63 \\
Ours-SoM-QwenVL-8B                
& 56.57 & 49.19 & 50.08 
& 57.33 & 49.63 & 50.30 
& 45.06 \\
Ours-SoM-QwenVL-32B               
& 59.65 & 57.46 & 56.77 
& 59.78 & 58.34 & 56.73 
& 52.42 \\
Ours-SoM-QwenVL-8B (Oracle)       
& \textbf{75.85} & \textbf{66.85} & \textbf{67.86} 
& \textbf{76.35} & \textbf{68.14} & \textbf{68.38} 
& \textbf{64.26} \\
\bottomrule
\end{tabular}
\end{adjustbox}
\caption{
Segmentation grounding performance on \textsc{ChartREG++}. 
We separate zero-shot methods from our synthetic-supervised two-stage setting. 
The Ours-SoM rows use a Mask2Former candidate generator trained with our synthesized chart masks, and therefore evaluate the utility of our mask synthesis pipeline rather than a zero-shot MLLM baseline. 
Gold-category and oracle rows are analysis settings.
}
\label{tab:seg_results}
\end{table*}

\subsection{Segmentation Grounding}
\label{sec:seg_grounding_res}

In this section, we evaluate segmentation grounding on \textsc{ChartREG++}. 
Importantly, we separate methods by their supervision setting. 
The Sa2VA and SAM3-SoM baselines are evaluated in a zero-shot setting, while our two-stage pipeline uses a chart-specific candidate generator trained with masks synthesized by our method. 
Thus, the latter is not intended as a fair zero-shot comparison against general-purpose MLLMs; instead, it evaluates whether our synthesized masks can \emph{\textbf{provide useful supervision for learning high-quality chart instance candidates and the ChartREG still leave room for improvement}}.

\begin{table*}[t]
\centering
\small
\setlength{\tabcolsep}{2pt}
\renewcommand{\arraystretch}{1.10}
\begin{adjustbox}{max width=\textwidth}
\begin{tabular}{@{}lccccccccccc@{}}
\toprule
& \multicolumn{3}{c}{\textbf{HBar}} 
& \multicolumn{3}{c}{\textbf{VBar}} 
& \multicolumn{3}{c}{\textbf{Pie}} 
& \multicolumn{2}{c}{\textbf{Line+Marker}} \\
\cmidrule(lr){2-4}\cmidrule(lr){5-7}\cmidrule(lr){8-10}\cmidrule(l){11-12}
\textbf{Method} 
& \textbf{P} & \textbf{R} & \textbf{F1} 
& \textbf{P} & \textbf{R} & \textbf{F1} 
& \textbf{P} & \textbf{R} & \textbf{F1} 
& \textbf{Det. rate} & \textbf{Avg. area} \\
\midrule
\multicolumn{12}{@{}l}{\textbf{Backbone: Qwen3VL-8B}} \\
ChartLens-SOM   
& 78.79 & 78.79 & 78.79 
& \textbf{92.21} & \textbf{91.34} & \textbf{91.56} 
& 88.38 & 85.83 & 86.37 
& 22.55 & 84.54 \\
SAM3-SOM        
& 39.39 & 39.39 & 39.39 
& 76.30 & 76.30 & 76.30 
& 84.79 & 82.60 & 83.18 
& 3.92 & \textbf{0.02} \\
Ours-SOM        
& \textbf{90.91} & \textbf{90.91} & \textbf{90.91} 
& 85.71 & 85.71 & 85.71 
& \textbf{93.55} & \textbf{91.79} & \textbf{92.03} 
& \textbf{80.39} & 1.38 \\
\midrule
\multicolumn{12}{@{}l}{\textbf{Backbone: Qwen3VL-32B}} \\
ChartLens-SOM   
& 78.79 & 78.79 & 78.79 
& \textbf{93.51} & \textbf{92.64} & \textbf{92.86} 
& 87.96 & 87.63 & 87.50 
& 22.55 & 40.43 \\
SAM3-SOM        
& 36.36 & 36.36 & 36.36 
& 78.90 & 78.90 & 78.90 
& 85.57 & 84.76 & 84.97 
& 3.92 & \textbf{0.02} \\
Ours-SOM        
& \textbf{90.91} & \textbf{90.91} & \textbf{90.91} 
& 88.74 & 88.74 & 88.74 
& \textbf{94.74} & \textbf{94.51} & \textbf{94.52} 
& \textbf{75.49} & 1.43 \\
\midrule
\multicolumn{12}{@{}l}{\textbf{Backbone: InternVL-8B}} \\
ChartLens-SOM   
& 78.79 & 78.79 & 78.79 
& 84.42 & 84.42 & 84.42 
& 85.76 & 83.75 & 84.07 
& 7.84 & 13.06 \\
SAM3-SOM        
& 36.36 & 36.36 & 36.36 
& 73.70 & 72.84 & 73.05 
& 80.00 & 78.04 & 78.52 
& 0.98 & \textbf{0.02} \\
Ours-SOM        
& \textbf{81.82} & \textbf{81.82} & \textbf{81.82} 
& \textbf{88.31} & \textbf{87.45} & \textbf{87.66} 
& \textbf{92.15} & \textbf{90.22} & \textbf{90.43} 
& \textbf{48.04} & 1.03 \\
\bottomrule
\end{tabular}
\end{adjustbox}
\caption{Performance on ChartLens-ChartQA real-world charts.}
\label{tab:chartlens_results}
\end{table*}
\subsubsection{Models}
We evaluate two segmentation grounding paradigms. 
\textbf{E2E-based} methods use the Sa2VA~\cite{yuan2025sa2vamarryingsam2llava} family, which directly predicts dense masks from the input image and referring expression. 
Since these models do not output separated chart instances, we only report $\mathrm{mIoU}_{\text{union}}$ for them. \textbf{Two-stage-based} methods decouple candidate generation and language alignment. 
The first stage generates instance mask candidates, and the second stage uses Set-of-Mark (SoM) prompting~\cite{yang2023setofmarkpromptingunleashesextraordinary} to let an MLLM select masks matching the referring expression. 
We include three candidate-generation settings. 
(1) \textbf{SAM3-SoM} uses the SAM3 segment-everything pipeline~\cite{carion2025sam3segmentconcepts} with ChartAgent-style post-processing, including area thresholding, mask-level NMS, removing composite masks, and filtering masks dominated by white background pixels. 
(2) \textbf{Ours-SoM (all segs)} uses a Mask2Former~\cite{cheng2022maskedattentionmasktransformeruniversal} model fine-tuned with masks synthesized by our pipeline. 
The training data includes 40K randomly sampled ChartCoder~\cite{zhao-etal-2025-chartcoder} examples and the ECDBench~\cite{yang2025effectivetrainingdatasynthesis} training split. 
(3) \textbf{Ours-SoM (gold-category segs)} further keeps only candidates whose category matches the target category of the referring expression. 
This is an analysis setting that assumes access to the target category. 
We also report \textbf{Ours-SoM (Oracle)}, where predicted candidates are replaced by ground-truth masks of the target category, to estimate the upper bound of the SoM+MLLM alignment stage. More details are shown in appendix~\ref{subsec:seg_grounding_details}. We use greedy sampling for all models.

\subsubsection{Evaluation metrics}
For methods that output separated instance masks, we perform one-to-one matching between predicted and ground-truth masks using the Hungarian algorithm. 
For common element types, we use mask IoU as the similarity score and count a match as true positive if $\mathrm{IoU}\ge 0.5$. 
For thin-line elements, including Errorbar, BoxMedianLine, Line\_with\_Points, and Polar\_Line\_with\_Points, we use Boundary IoU averaged over $d\in\{1,2,4,8\}$.\footnote{The line-width distribution in our dataset mainly falls in the range of 1--15 pixels.}
We compute macro Precision, Recall, and F1 from TP/FP/FN. 
We also report $\mathrm{mIoU}_{\text{union}}$, the sample-wise IoU between the union of predicted masks and the union of ground-truth masks, which measures overall region coverage.

\subsubsection{Results}
As shown in Table~\ref{tab:seg_results}, zero-shot segmentation grounding remains difficult on \textsc{ChartREG++}. 
Both Sa2VA and SAM3-SoM obtain limited performance, showing that general-purpose segmentation grounding methods struggle with fine-grained chart elements and multi-target references.

Our synthetic-supervised \textbf{Ours-SoM (all segs)} achieves much higher scores than SAM3-SoM. 
This result should not be read as a fair zero-shot comparison, since our first-stage Mask2Former is trained with synthesized chart masks. 
Rather, it shows that our synthesis pipeline provides effective supervision for learning cleaner chart instance candidates, and that candidate quality is a key bottleneck in two-stage grounding. 
The gold-category setting further improves performance by reducing the candidate search space, while the oracle row still leaves a clear gap from perfect performance. 
This suggests that, beyond candidate generation, instance-level language alignment under diverse cues and multi-target expressions remains challenging. The remaining oracle gap also suggests that \textsc{ChartREG++} is far from saturated and leaves clear room for future improvement.

\subsection{Generalize to real-world Charts}
We evaluate whether our model generalizes to real-world charts by testing on the \textsc{ChartQA} subset in \textsc{ChartLens} \emph{\textbf{whose images are non-matplotlib}}.\footnote{We find that, for line charts, some point annotations in \textsc{ChartLens} are offset from the polyline markers; we correct them by moving the points onto the corresponding markers. We also rewrite all questions into referring-grounding expressions (e.g., what's the maximum value of the bars?'' $\rightarrow$ segment the bar with the maximum value'') and manually verify that each expression uniquely and precisely refers to the target objects. More details are shown in Appendix~\ref{subsec:chartlens_results}}
We follow the evaluation protocol of \textsc{ChartLens}.
For each image and each target category, we generate candidate masks using (i) the \textsc{ChartLens} rule-based mask generation method, (ii) \textsc{SAM3} with a segment-everything pipeline, and (iii) our method. 
We then apply \textsc{Set-of-Mark} prompting to these candidates and evaluate on different base MLLMs.
Results in Table~\ref{tab:chartlens_results} show that our method achieves better performance than the baselines on most categories; in particular, on line-related categories, it not only recalls polyline points effectively but also yields negligible area coverage.

\section{Conclusion}
We introduce ChartREG++, a chart-domain referring expression grounding benchmark with diverse cues, multi-target queries, broad element coverage, and multiple localization formats, together with a mask synthesis that generates controllable multi-granularity fine-grained label instance masks. Experiments show consistent gains and transfer to real-world non-matplotlib charts on ChartLens. We hope ChartREG++ will promote further progress toward more verifiable chart understanding.

\section*{Limitations}
\textsc{ChartREG++} has several directions for further extension.
First, our current synthesis pipeline mainly focuses on Matplotlib-based static charts, which allows us to obtain accurate and scalable mask annotations. 
Future work may extend the same idea to more visualization libraries and interactive chart settings. 
Second, although \textsc{ChartREG++} covers diverse referring cues and multi-target cases, it still focuses on commonly used chart elements and referring patterns.  More domain-specific charts and more user-written expressions could be added in future versions.

\bibliography{custom}

\appendix

\section{Appendix}
\label{sec:appendix}


\begin{table*}[t]
\centering
\normalsize
\setlength{\tabcolsep}{3.5pt}
\renewcommand{\arraystretch}{1.15}
\begin{adjustbox}{max width=\textwidth}
\begin{tabular}{@{}llll@{}}
\toprule
\textbf{Name} 
& \textbf{Output format} 
& \textbf{Task} 
& \textbf{Ref. clues} \\
\midrule
InfoDet        
& BBox                  
& Element localization        
& -- \\
RefChartQA     
& BBox                  
& Attribution QA              
& Textual/Data (only Data-Rank clue) ($>$90\%) \\
ChartLens      
& BBox                  
& Attribution QA              
& Textual/Data (only Data-Rank clue) ($>$90\%) \\
ChartRef       
& BBox                  
& Referring grounding         
& Textual \\
DOGR (Chart)   
& BBox                  
& Attribution QA              
& -- \\
\textbf{ChartREG++}  
& \textbf{Point/Bbox/Seg.} 
& \textbf{Referring grounding} 
& \textbf{Textual/Data/Visual} \\
\bottomrule
\end{tabular}
\end{adjustbox}
\caption{Comparison of our benchmark with existing chart visual grounding-related benchmarks.}
\label{tab:benchmark_comparison_2}
\end{table*}


\begin{table*}[t]
\centering
\normalsize
\setlength{\tabcolsep}{3.5pt}
\renewcommand{\arraystretch}{1.15}
\begin{adjustbox}{max width=\textwidth}
\begin{tabular}{@{}lccccl@{}}
\toprule
\textbf{Name} 
& \textbf{\#Elem.} 
& \textbf{Avg. targets} 
& \textbf{\#Images} 
& \textbf{\#Samples} 
& \textbf{Chart source} \\
\midrule
InfoDet \cite{zhu2026infodet}        
& 26 & --  & 100k   & --    & Synthetic + real charts \\
RefChartQA \cite{vogel2025refchartqagroundingvisualanswer}     
& 3  & 2.1 & $<$1.6k & 11.7k & ChartQA \\
ChartLens \cite{suri-etal-2025-chartlens}      
& 3  & 1.8 & 1,148 & 1.25k & ChartQA \\
ChartRef \cite{tjandrasuwita2025chartref}       
& 11 & 1.0 & 1,141 & 38.9k & ChartMimic \\
DOGR (Chart) \cite{zhou2025dogrversatilevisualdocument} 
& 1  & --  & 600   & 0.6k  & ChartQA \\
\textbf{ChartREG++}  
& \textbf{18} & \textbf{9.7} & \textbf{850} & \textbf{3.4k} 
& \textbf{ChartMimic + ECDBench} \\
\bottomrule
\end{tabular}
\end{adjustbox}
\caption{Comparison of our benchmark with existing chart visual grounding-related benchmarks. We report test-split statistics for RefChartQA, ChartLens , DOGR (Chart) and ChartREG++. We report train and test-split statistics for InfoDet, ChartRef.}
\label{tab:benchmark_comparison_1}
\end{table*}
\subsection{Related Work}
\label{sec:related_work}

\subsubsection{Chart Understanding}
\label{subsec:chart_understanding}
Recent Chart understanding and generation research mainly covers chart question answering, structured parsing , and chart summarization chart-to-text. Representative benchmarks such as ChartQA \cite{masry-etal-2022-chartqa}, CharXiv \cite{wang2024charxiv}, ChartMusum\cite{tang2025chartmuseum} emphasize visual and logical reasoning on real-world charts, These efforts have advanced multimodal chart understanding, but they also expose a key limitation: when models output only a textual answer or description, the result is hard for users to directly verify against the chart.

\subsubsection{Chart Visual Grounding}
\label{subsec:chart_visual_grounding}
Recent work chart visual grounding-related works can be categorized into four problem settings. Some approaches embed pointing or bounding-box supervision into multimodal reasoning traces, aiming to improve chart reasoning and reduce hallucinations (e.g., Point-RFT \cite{ni2025pointrft}, ChartPoint \cite{xu2025chartpointguidingmllmsgrounding}). Another line of benchmarks evaluates answer-attribution by requiring models to output both the final answer and the corresponding evidence region, and further studies finer-grained attribution with segmentation and Set-of-Marks prompting (e.g., RefChartQA \cite{vogel2025refchartqagroundingvisualanswer}, ChartLens \cite{suri-etal-2025-chartlens},DOGR \cite{zhou2025dogrversatilevisualdocument}). The third line focuses on chart-domain referring expression grounding, where the model directly maps an expression to its referenced region without producing a textual answer (e.g., ChartRef \cite{tjandrasuwita2025chartref}). The fourth line is detection-oriented datasets such as InfoDet \cite{zhu2026infodet} which provide large-scale bounding-box annotations for chart infographic elements. However, from a task-design perspective, existing work rarely evaluates referring grounding under the joint setting of multiple referred targets, multiple cue types, and multiple localization formats. From a data-synthesis perspective, current synthesis pipelines are also limited in generating chart instance masks with multi-granularity definitions and finer semantic labels for chart elements.

\begin{table*}[t]
\centering
\caption{Target element type categories and their prompts used in instance segmentation for GroundingSAM2 and SAM3}
\label{tab:chart_element_prompts}
\scriptsize
\setlength{\tabcolsep}{5pt}
\renewcommand{\arraystretch}{1.12}
\begin{tabularx}{\textwidth}{>{\raggedright\arraybackslash}p{0.22\textwidth} >{\raggedright\arraybackslash}X}
\toprule
\textbf{Category} & \textbf{Prompt Variants} \\
\midrule
BoxMedianLine & boxplot median line; box plot median line; median line in boxplot; median line (box plot) \\

BoxPlot\_Boxpatch & boxplot box; box plot box (IQR); interquartile range box; box-and-whisker box; IQR rectangle (boxplot) \\

Fill\_between\_density & filled band under line series; filled band under line series; filled area between line series and axis \\

Fill & filled closed area by line series; filled region enclosed by polar plot line \\

Hist & histogram bar; histogram bin bar; histogram bin; binned bar (histogram) \\

Stackplot\_area & stacked area chart layer; stacked area layer; area stack layer; stacked area band \\

Full Box & box along with its whisker, median line and caps \\

PolarLine\_withPoints & polar line series; single polar line plot (data series); radial line series; line in polar coordinates; polar polyline data series; polar chart line series \\

Line\_withPoints & line plot series; data line series; connected line series (x-y); single cartesian line plot; polyline data series; line plot (data series) \\

LinePoints & line chart marker; line plot marker; marker on line series; data point marker on line; line series point marker \\

PolarLinePoints & polar line plot marker; marker on polar line; polar line series markers \\

Scatter & scatter marker point \\

HBar & horizontal bar (chart); horizontal bar chart bar; horizontal bar segment; bar in horizontal bar chart; hbar (data bar) \\

VBar & vertical bar (chart); vertical bar chart bar; column bar (chart); bar in vertical bar; vbar (data bar) \\

PolarVBar & polar bar sector; radial bar sector; polar chart bar sector; circular bar sector; polar bar wedge \\

ErrorBar & error bar \\

PieSector & pie slice; pie chart slice; pie sector; pie wedge; donut slice; donut chart slice; annular wedge; annular sector; ring sector \\

Treemap & treemap sector; treemap rectangle \\
\bottomrule
\end{tabularx}
\end{table*}
\begin{figure*}[t]
  \centering
  \begin{minipage}[t]{0.33\textwidth}
    \centering
    \begin{tabular}[t]{@{}c@{}}
      \includegraphics[width=\linewidth]{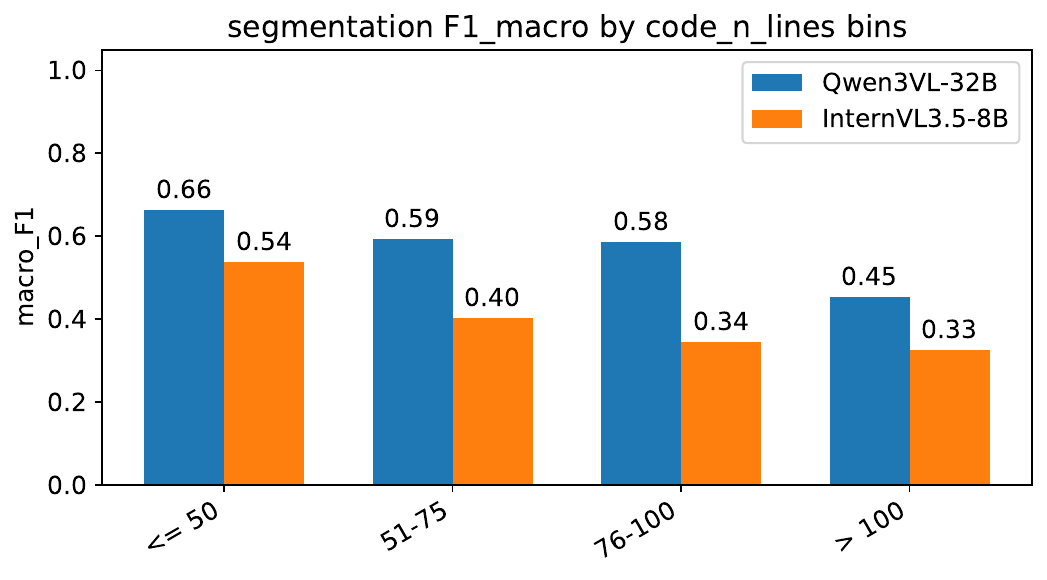}
    \end{tabular}
  \end{minipage}\hfill
  \begin{minipage}[t]{0.33\textwidth}
    \centering
    \begin{tabular}[t]{@{}c@{}}
      \includegraphics[width=\linewidth]{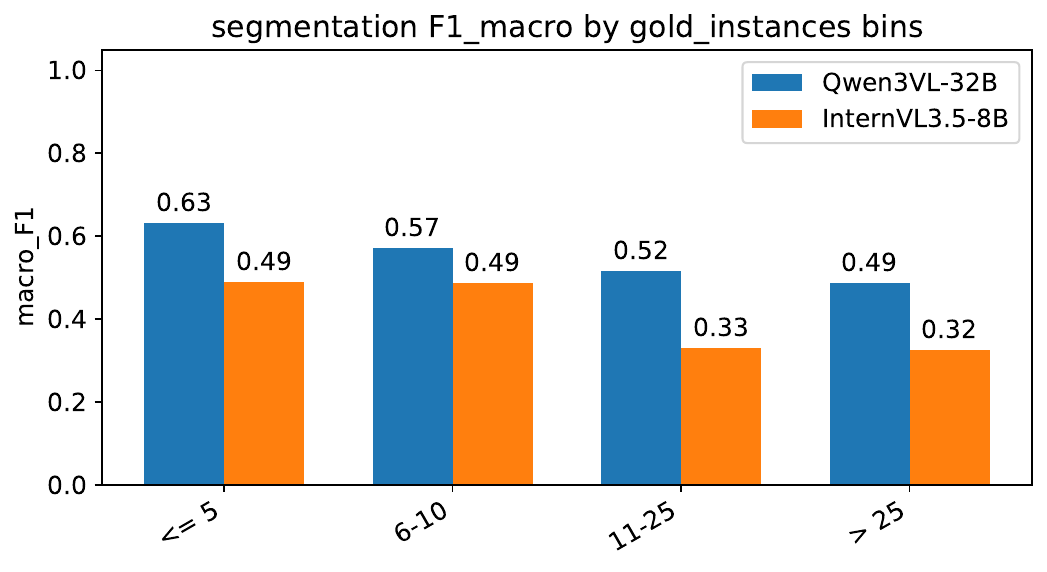}
    \end{tabular}
  \end{minipage}\hfill
  \begin{minipage}[t]{0.33\textwidth}
    \centering
    \begin{tabular}[t]{@{}c@{}}
      \includegraphics[width=\linewidth]{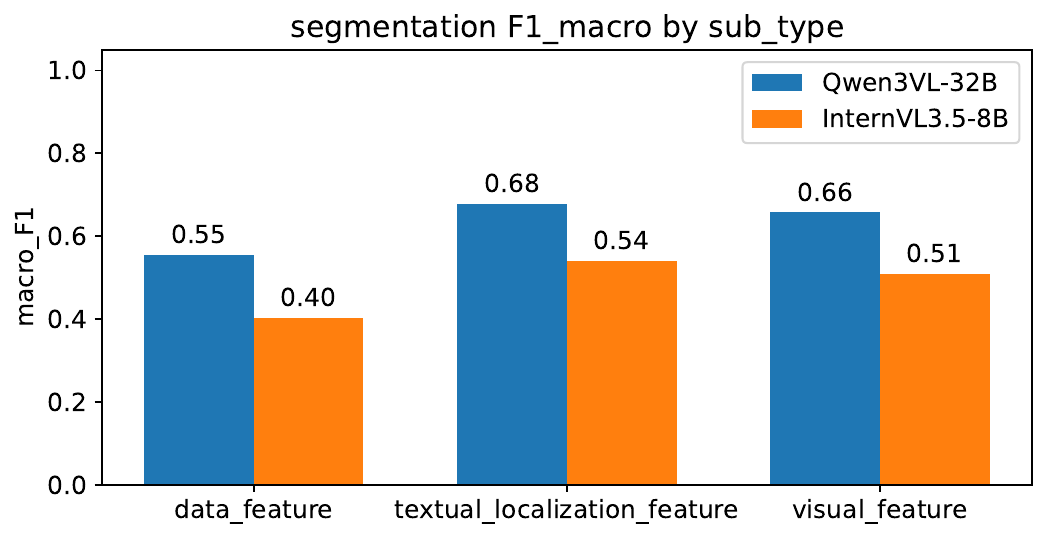}
    \end{tabular}
  \end{minipage}

  \par\noindent 

  \includegraphics[width=\textwidth]{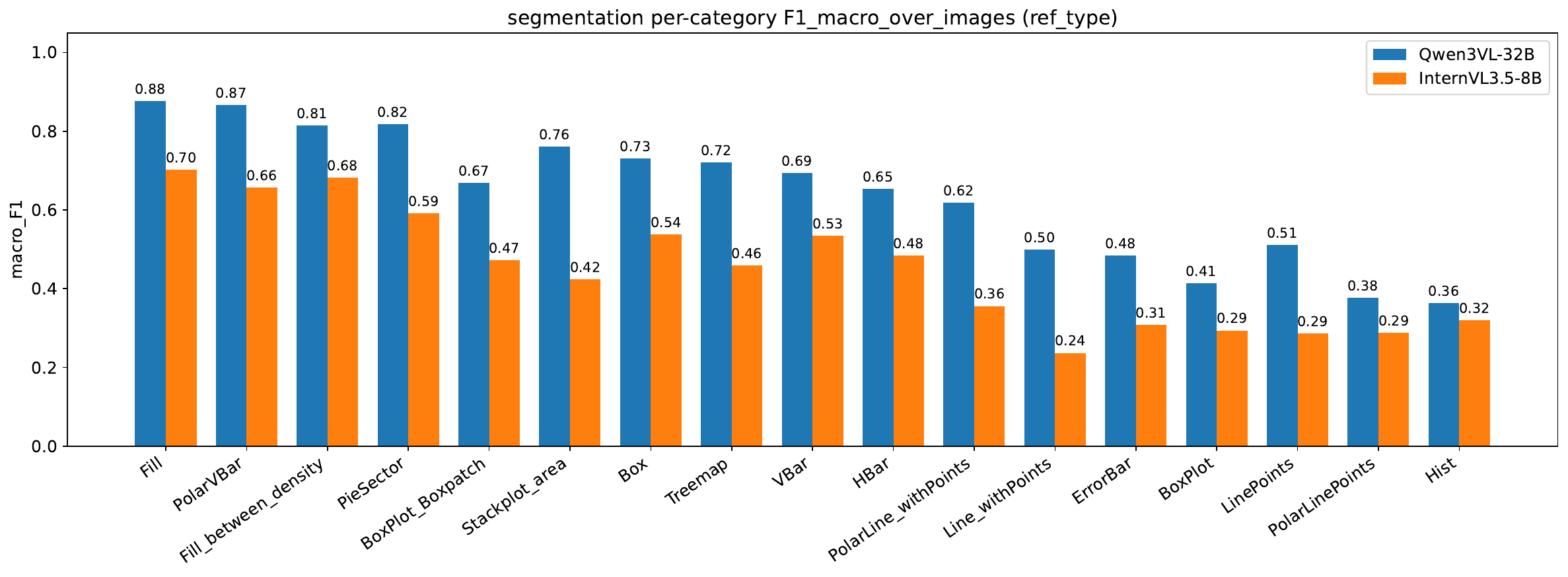}

  \caption{%
  Break down analysis results.
  }
  \label{fig:merged_overview}
\end{figure*}
\subsection{Comparison with Existing Benchmarks}

We compare ChartREG++ with existing works in Table~\ref{tab:benchmark_comparison_2} and Table~\ref{tab:benchmark_comparison_1}.

\subsection{More Experimental Details and Results}
\label{sec:more_exp_details}
\subsubsection{Segmentation grounding details}
\label{subsec:seg_grounding_details}
For Sa2VA, we use the official default settings for inference. For candidate mask generation in \textbf{SAM3-Seg-everything}, we formulate the process as a \textbf{class-agnostic automatic instance proposal} problem rather than explicit text-conditioned segmentation. Specifically, for each image, we use automatic mask generation with regularly sampled points over the full image for prompt-free segmentation. We set the number of sampled points per batch to \textbf{64}, the total number of points to \textbf{1024}, and the predicted mask quality threshold to \textbf{0.88} to filter out low-quality candidates. Since this process often produces many redundant, overly coarse, or background-driven masks, we further follow ChartAgent and apply a set of conservative post-processing rules. We first remove very small regions with an area smaller than \textbf{10} pixels. We then deduplicate highly overlapping candidates using a mask IoU threshold of \textbf{0.9}. For coarse composite masks that are largely covered by the union of other candidates, we discard them using a coverage threshold of \textbf{0.98}. Finally, to suppress false candidates caused by the large white background in chart images, we remove masks whose internal white-pixel ratio exceeds \textbf{0.95}, where a white pixel is defined as having all three RGB channels no smaller than \textbf{245}.

\subsubsection{Instance Segmentation results and details}


\begin{table}[t]
\centering
\footnotesize
\setlength{\tabcolsep}{4pt}
\renewcommand{\arraystretch}{1.12}
\begin{adjustbox}{max width=\columnwidth}
\begin{tabular}{@{}lcccccc@{}}
\toprule
\textbf{Model} 
& \textbf{mAP} 
& \textbf{mAP@50} 
& \textbf{mAP@75} 
& \textbf{mAP$_s$} 
& \textbf{mAP$_m$} 
& \textbf{mAP$_l$} \\
\midrule
GroundingSAM2  
& 0.58 & 0.81 & 0.56 & 0.02 & 0.11 & 0.85 \\
SAM3-Detection 
& 10.25 & 11.72 & 10.99 & 1.33 & 7.99 & 23.28 \\
\textbf{Ours-Detection} 
& \textbf{51.59} & \textbf{69.34} & \textbf{52.58} 
& \textbf{20.25} & \textbf{47.03} & \textbf{74.73} \\
\bottomrule
\end{tabular}
\end{adjustbox}
\caption{Instance segmentation performance in COCO-style mAP.}
\label{tab:instance_segmentation_res}
\end{table}

\label{subsec:instance_seg_details}
We compare our fine-tuned \textsc{Mask2Former} with two off-the-shelf instance segmentation baselines, \textbf{GroundedSAM2} \cite{ren2024groundedsamassemblingopenworld} and \textbf{SAM3} \cite{carion2025sam3segmentconcepts}, on \textsc{ChartREG++} images.
To run these baselines, for each target category we manually design multiple short phrase prompts; for each image we only execute prompts for the categories that appear in the image's ground-truth annotations, merge masks from different prompts of the same category by union, and then remove duplicates. We compute standard COCO-style instance segmentation metrics using the official evaluation protocol,and compare them with our model. Results in Table~\ref{tab:instance_segmentation_res} show that our model outperforms these general-purpose baselines overall, while performance on small instances still leaves clear room for improvement.

For GroundingSAM2 and SAM3, we predefine, for each category, a set of natural language prompts that match the semantics of the corresponding chart element as shown in Table~\ref{tab:chart_element_prompts}. When a category includes multiple synonymous prompts, we run inference for each prompt independently and merge all outputs into the candidate set of that category. During inference, we use a confidence threshold of \textbf{0.5} to filter out low-quality candidates. We then perform category-wise deduplication based on mask IoU with a threshold of \textbf{0.9}, in order to suppress duplicate instances triggered by different synonymous prompts. We train Mask2Former using the training split of ECDBench and 40K randomly sampled samples from ChartCoder. Since ECDBench provides separate training and evaluation splits, we use only its training split for model training. ChartCoder is a training dataset for the chart-to-code task. Therefore, this setting does not introduce data leakage.

\subsubsection{Break down analysis results}
\label{subsec:breakdown_analysis}

We conduct more fine-grained qunatitative analysis with different subsets of our benchmark using our model in Sec~\ref{sec:seg_grounding_res}.

\paragraph{Effect of chart complexity.}
We measure chart complexity by the plotting-code length. As shown in Figure~\ref{fig:merged_overview} top left, performance decreases as code length increases, indicating that grounding in more complex charts is still difficult.
\paragraph{Effect of the number of referred targets.}
As shown in Figure~\ref{fig:merged_overview} top middle, performance consistently drops as the number of referred targets increases, suggesting that \textbf{multi-target} grounding remains challenging.
\paragraph{Effect of primary clue categories.}
As shown in Figure~\ref{fig:merged_overview} top right, the model performs well on \textbf{textual/location} and \textbf{visual} cues, indicating strong performance when the grounding mainly relies on perceptual signals. In contrast, \textbf{data} cues, which often require comparisons or simple reasoning over values, still leave substantial room for improvement.
\paragraph{Effect of referred target element types.}
As shown in Figure~\ref{fig:merged_overview} bottom, performance varies significantly across referred element types. Overall, visually dense targets (e.g., \texttt{Hist}) and small-area targets (e.g., \texttt{Line}) still have large headroom.

\subsubsection{Generalize to real-world details (ChartLens-ChartQA)}
\label{subsec:chartlens_results}
\begin{figure*}[t]
  \centering
  \includegraphics[width=\textwidth]{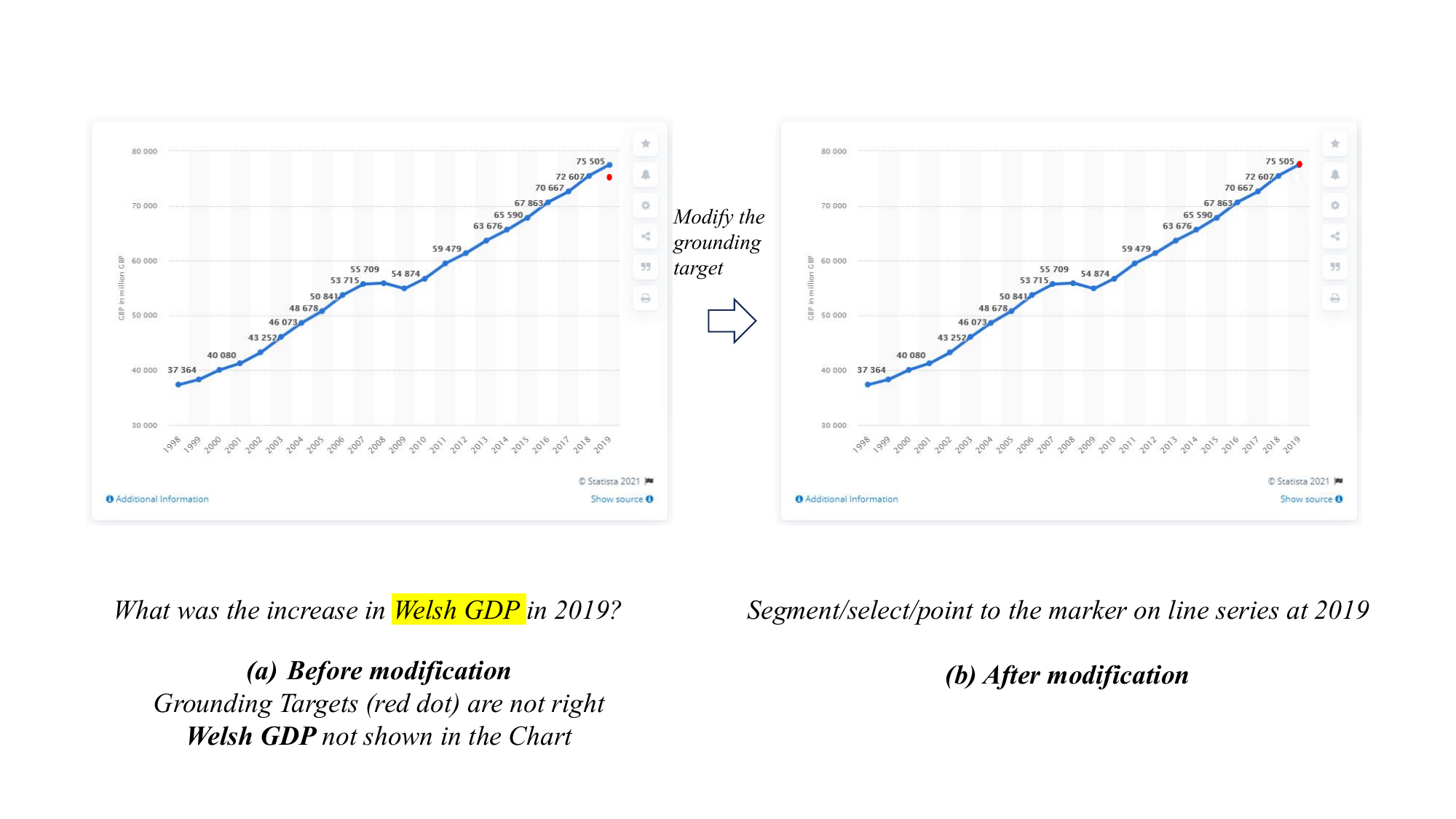}
  \caption{chartlens modification example}
  \label{fig:chartlens_modification}
\end{figure*}

We find that the original ChartLens-ChartQA test set contains two types of annotation errors. First, for line charts, the grounding targets do not match the actual targets required by the question. As shown in Figure~\ref{fig:chartlens_modification} (left), the grounding annotations in ChartLens-ChartQA do not align with the targets needed for answering the question. Second, some questions contain information that does not appear in the chart. To correct these errors and at the same time make the evaluation setting closer to our task, we revise the dataset format as illustrated in Figure~\ref{fig:chartlens_modification}. Specifically, we first correct the grounding targets in line charts by moving them to the markers of the target line series, as shown by the red points in the figure. We then revise the questions so that they no longer contain information absent from the chart, and convert them into referring expressions that can uniquely and precisely identify the intended targets. After these corrections, we conduct an additional round of manual verification to ensure that the final data no longer contain the above issues.

\subsubsection{Case study}
\begin{figure*}[t]
  \centering
  \includegraphics[width=\textwidth]{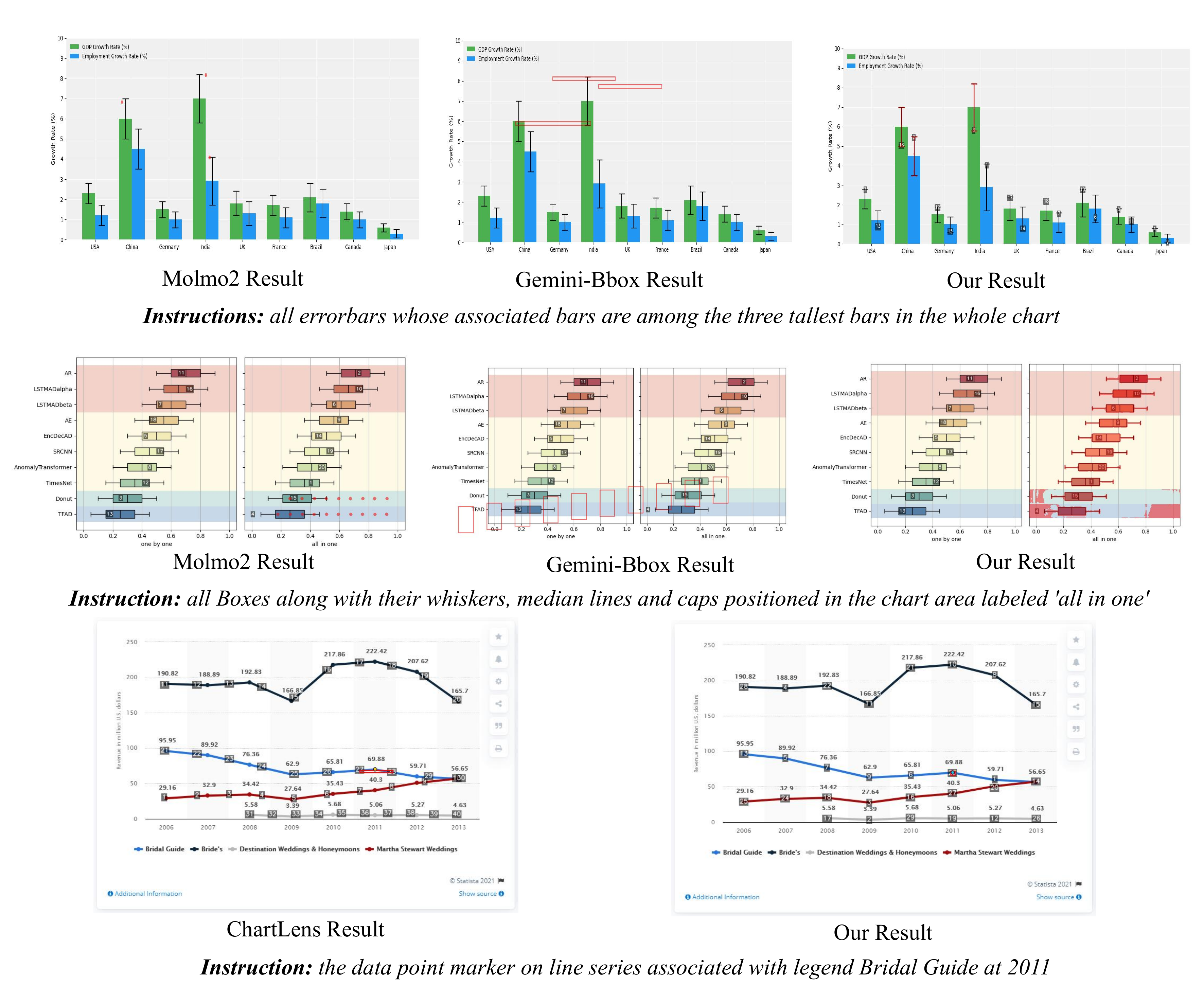}
  \caption{Qualitative cases between our method and existing methods}
  \label{fig:episode_overview}
\end{figure*}

As shown in Figure~\ref{fig:episode_overview} top, while \textsc{Molmo} can point near the error bars, it fails to correctly satisfy the constraint ``the highest three bars''. In addition, the Gemini-based bbox output is still inaccurate for the geometry of error bars, whereas our method localizes them correctly.

As shown in Figure~\ref{fig:episode_overview} (middle), we ask models to localize the referred target at the granularity of a \emph{complete box} in a box plot. \textsc{Molmo} produces points far from the target, suggesting it may not reliably capture this granularity definition. Gemini predicts the correct number of targets but with noticeable localization errors, indicating weaker execution from definition to precise position. Benefiting from our multi-granularity mask synthesis, our framework can provide candidates at this granularity and thus select the correct targets.

As shown in Figure~\ref{fig:episode_overview} (bottom), the \textsc{ChartLens} pipeline asks the MLLM to select \emph{two marked points} given by Lineformer on the polyline as the corners of a bounding box so that the box covers the target point. This requires an extra step of \emph{imagining/predicting} which point pair will form a covering box, which can fail even when the selected points are close to the target. In contrast, our method directly provides \emph{candidate point instances} (as masks) on the polyline, therefore the MLLM can select the target \emph{directly from these candidates} via SoM, without reasoning about which points should compose a covering bounding box, thereby reducing the attribution difficulty.

\subsection{More ChartREG++ Benchmark Details}

\label{sec:benchmark_details}
\begin{figure*}[t]
  \centering
  \begin{minipage}[t]{0.32\textwidth}
    \centering
    \includegraphics[width=\linewidth]{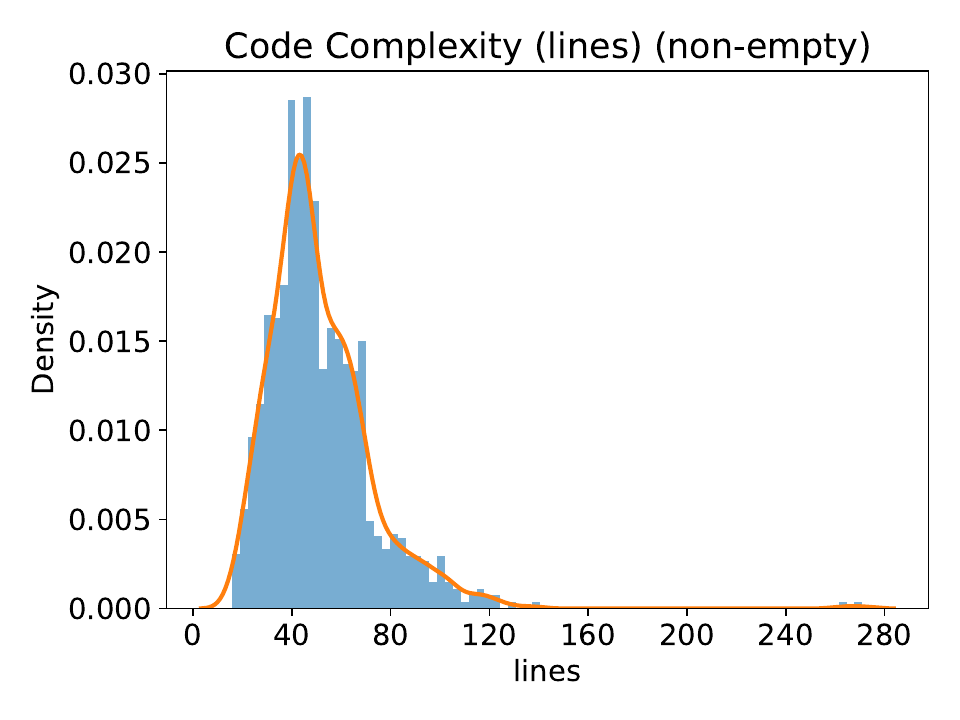}
  \end{minipage}\hfill
  \begin{minipage}[t]{0.32\textwidth}
    \centering
    \includegraphics[width=\linewidth]{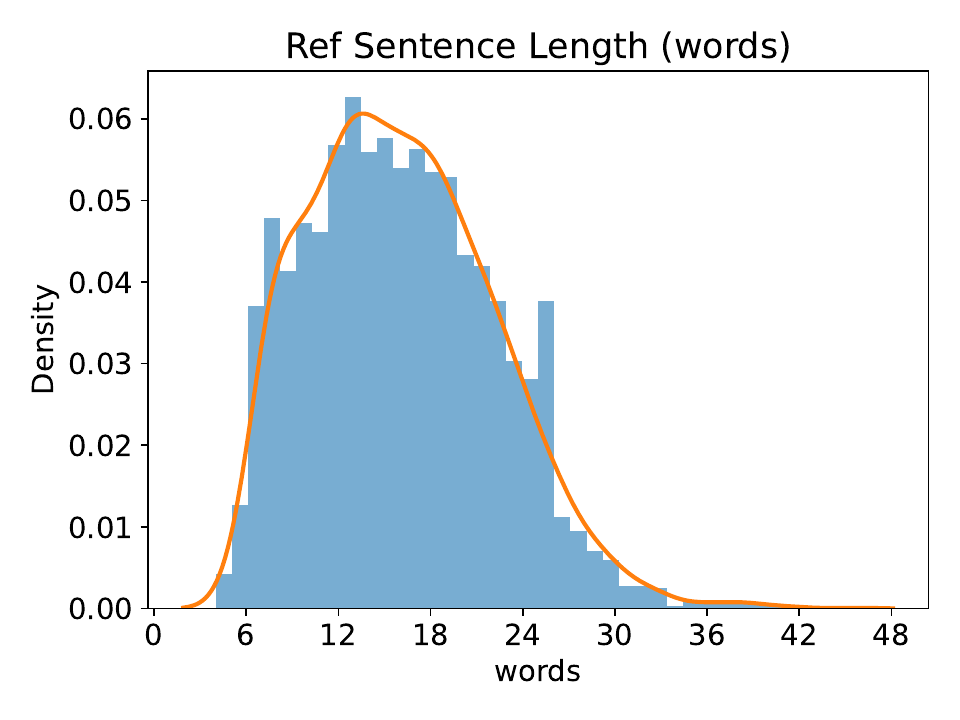}
  \end{minipage}\hfill
  \begin{minipage}[t]{0.32\textwidth}
    \centering
    \includegraphics[width=\linewidth]{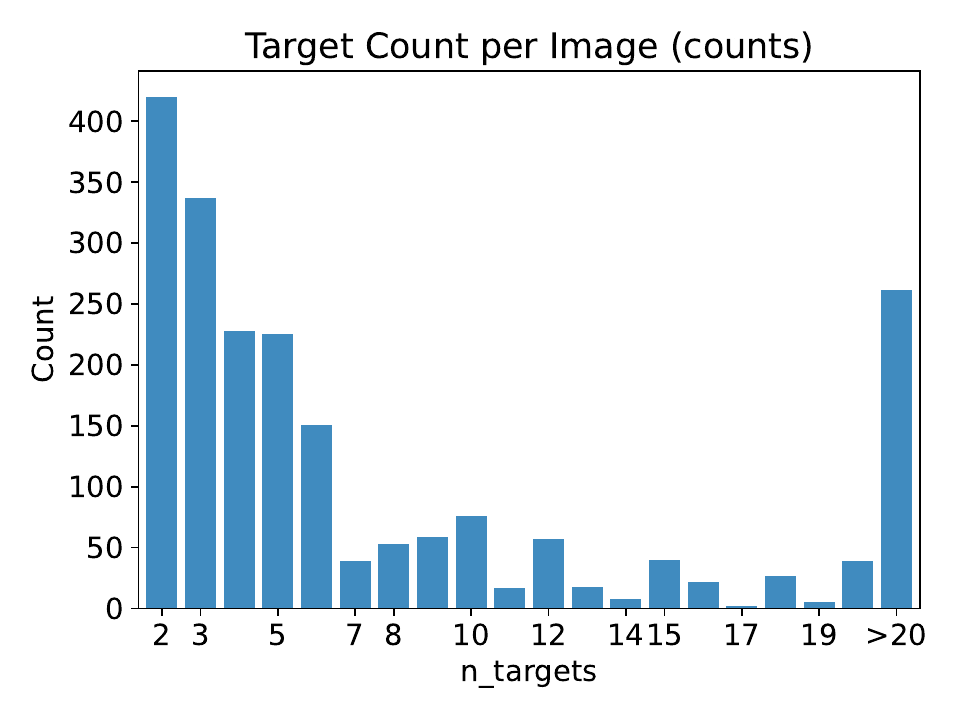}
  \end{minipage}

  \par\noindent 

  \begin{minipage}[t]{0.44\textwidth}
    \centering
    \includegraphics[width=\linewidth]{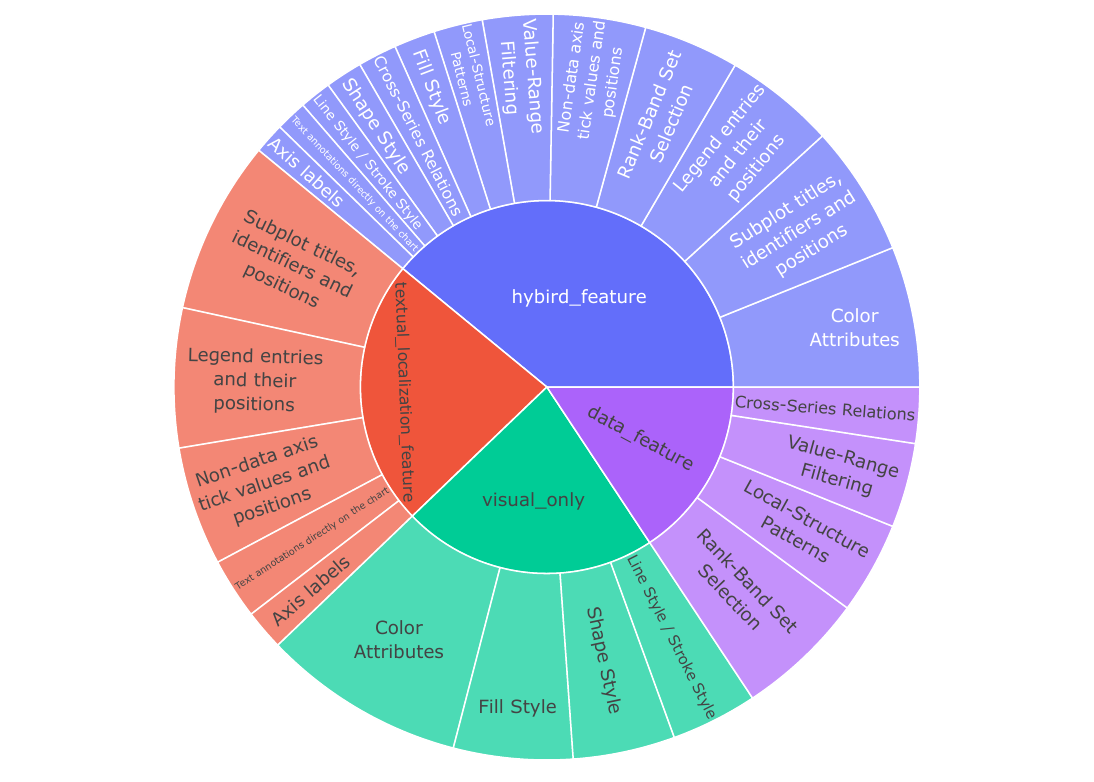}
  \end{minipage}\hfill
  \begin{minipage}[t]{0.54\textwidth}
    \centering
    \includegraphics[width=\linewidth]{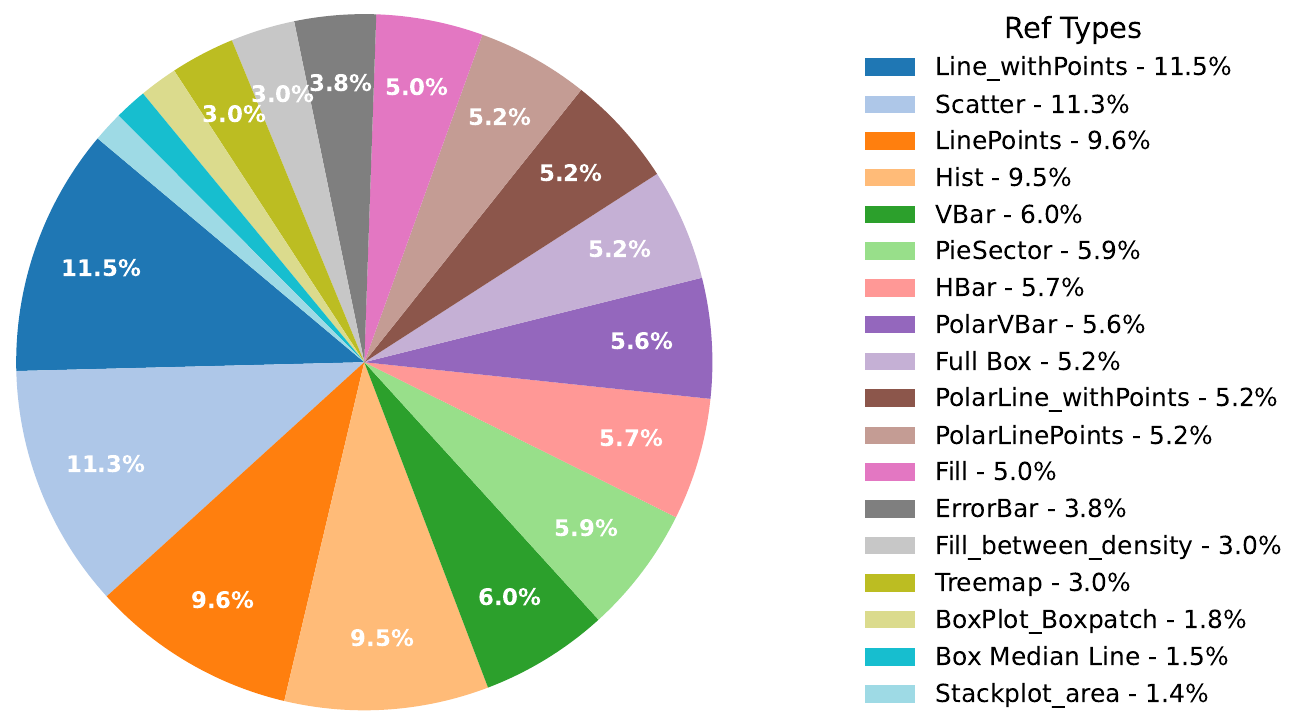}
  \end{minipage}

  \caption{Distributions of dataset complexity and taxonomy. Top: (left) target image complexity measured by the number of lines in the corresponding plotting code; (middle) complexity of referring expressions measured by sentence length; (right) distribution of the number of referred target instances per query (shown only for multi-target samples). Bottom: (left) distribution of referring cue types; (right) distribution of referred target element types.}
  \label{fig:ds_distribution_all}
\end{figure*}

\subsubsection{Discussion on benchmark scales}
\label{subsec:benchmark_scale_discussion}
CHARTREF mainly studies one-to-one phrase-to-box grounding of individual data elements, which enables large-scale construction but also introduces many structurally similar per-element instances. RefChartQA instead focuses on grounding the visual evidence needed to answer a QA query, inheriting its supervision structure from ChartQA/ChartQA-PoT and thus centering on answer justification rather than an explicit diversity of referring cues. Our benchmark follows a different design goal:  we prioritize coverage of grounding phenomena over repeated enumeration of similar instances.

\subsubsection{Data generation details}
\label{subsec:data_generation_details}
For each target element type, we use Gemini, DeepSeek, and GPT models to generate referring expressions. Table~\ref{tab:target_element_subjects} lists each target element type and the subject used when generating its referring expressions. The prompts used for referring expression generation are shown in Prompt~\ref{box:full_prompt_data_related} and Prompt~\ref{box:full_prompt_non_data_related}. For each synthesized referring expression, we use DeepSeek with the prompt in Prompt~\ref{box:full_prompt_ref_target_gen} to infer the referred targets, and then obtain the final mask corresponding to the referring expression using the method described in Sec~\ref{sec:method}. We then perform manual verification.

\begin{table*}[t]
\centering
\caption{Target element types and their subject forms in referring expressions.}
\label{tab:target_element_subjects}
\scriptsize
\setlength{\tabcolsep}{5pt}
\renewcommand{\arraystretch}{1.12}
\begin{tabularx}{\textwidth}{>{\raggedright\arraybackslash}p{0.27\textwidth} >{\raggedright\arraybackslash}X}
\toprule
\textbf{Target element type name} & \textbf{Subject in referring expression} \\
\midrule
VBar & vertical bar \\

PieSector & pie sector/ring sector \\

Line\_withPoints & line series along with its markers/plotted line series along with its markers \\

Scatter & scatter point/scatter marker \\

LinePoints & marker on plotted line series/marker on line series/plotted line series marker \\

Hist & histogram bar \\

HBar & horizontal bar \\

PolarVBar & polar bar sector \\

Full Box & Box along with its whisker, median line and caps \\

PolarLine\_with\_points & polar line series along with their markers/polar plot line series along with its marker/polar line series/polar plot line series \\

PolarLinePoints & marker on plotted polar line series/marker on polar line series/marker on line series \\

Fill & filled (parallelogram) region/filled polygon \\

ErrorBar & errorbar \\

Fill\_between\_density & filled band/filled region \\

Treemap & treemap sector/treemap rectangle \\

BoxPlot\_BoxPatch & box patch \\

Box Median Line & box median line \\

StackPlot\_area & stack area layer/stack series band/stack series area \\
\bottomrule
\end{tabularx}
\end{table*}

\subsection{Benchmark Statistics and Distribution}
We show the visualization of benchmark statistics and distribution in Figure~\ref{fig:ds_distribution_all}.

\subsubsection{Benchmark examples}
\label{subsec:benchmark_examples}
\begin{figure*}[t]
  \centering
  \includegraphics[width=\textwidth]{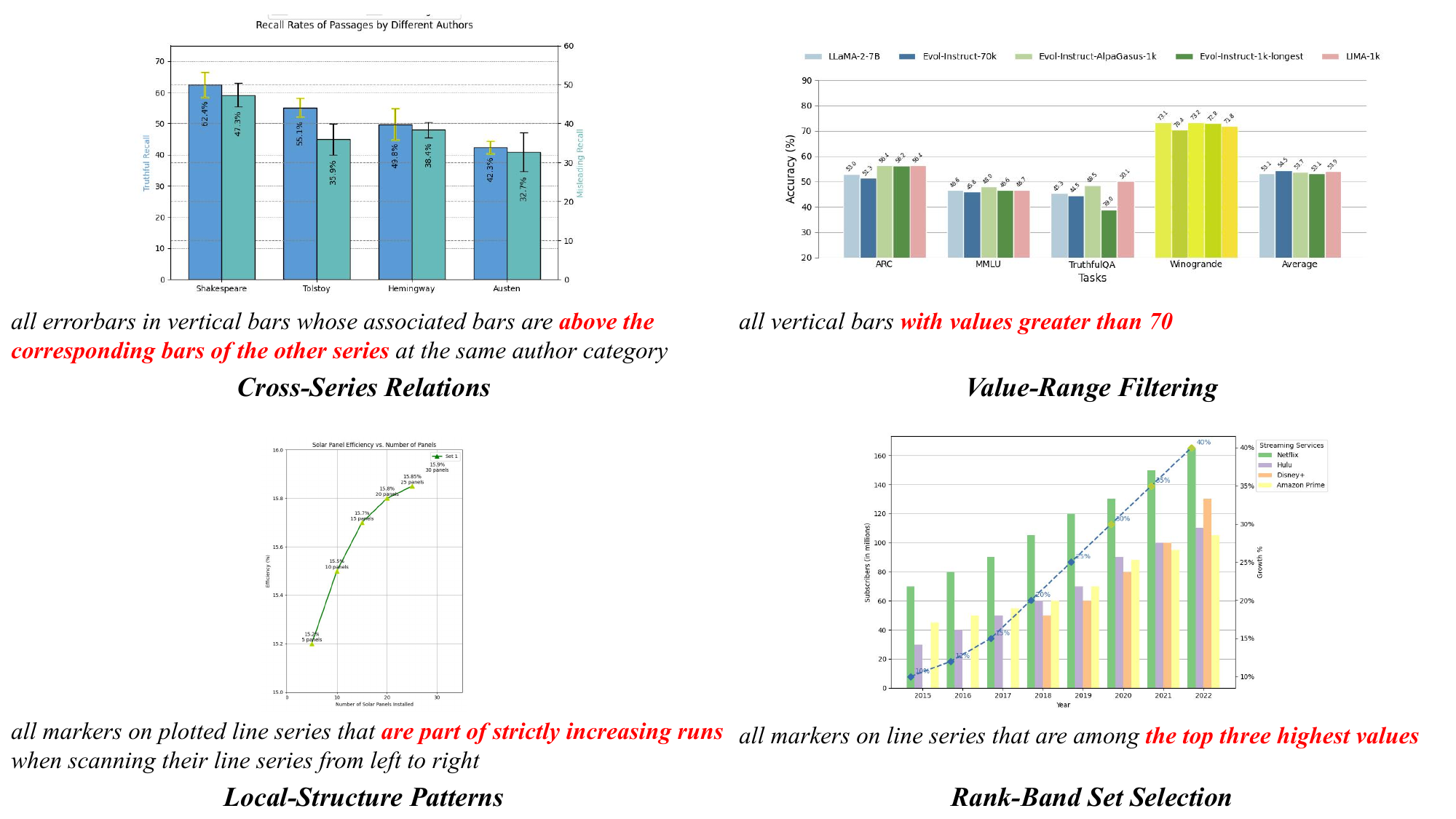}
  \caption{data referring clue example}
  \label{fig:data_referring_clue_example}
\end{figure*}

\begin{figure*}[h]
  \centering
  \includegraphics[width=\textwidth]{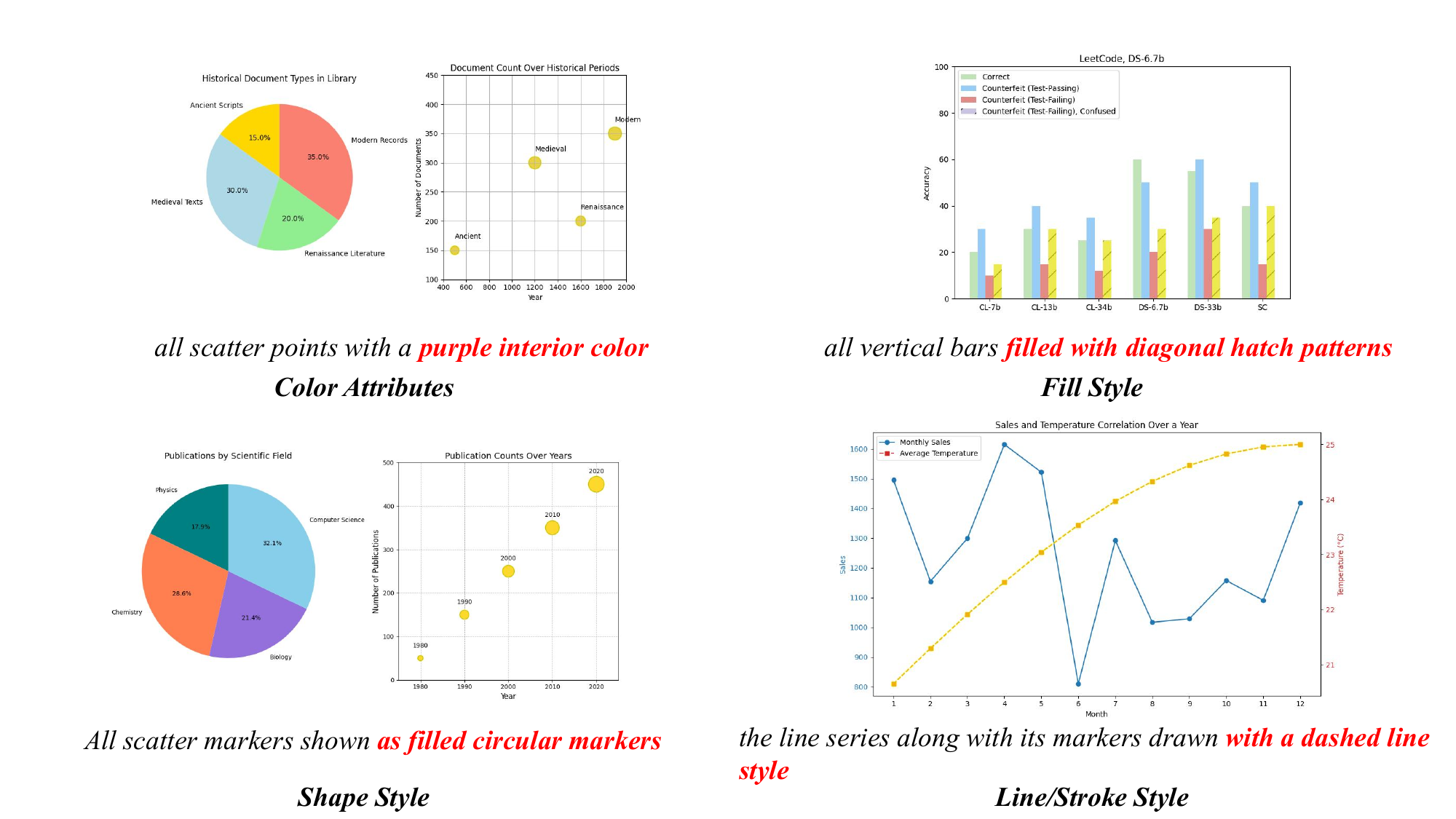}
  \caption{visual referring clue example}
  \label{fig:visual_referring_clue_example}
\end{figure*}

\begin{figure*}[h]
  \centering
  \includegraphics[width=\textwidth]{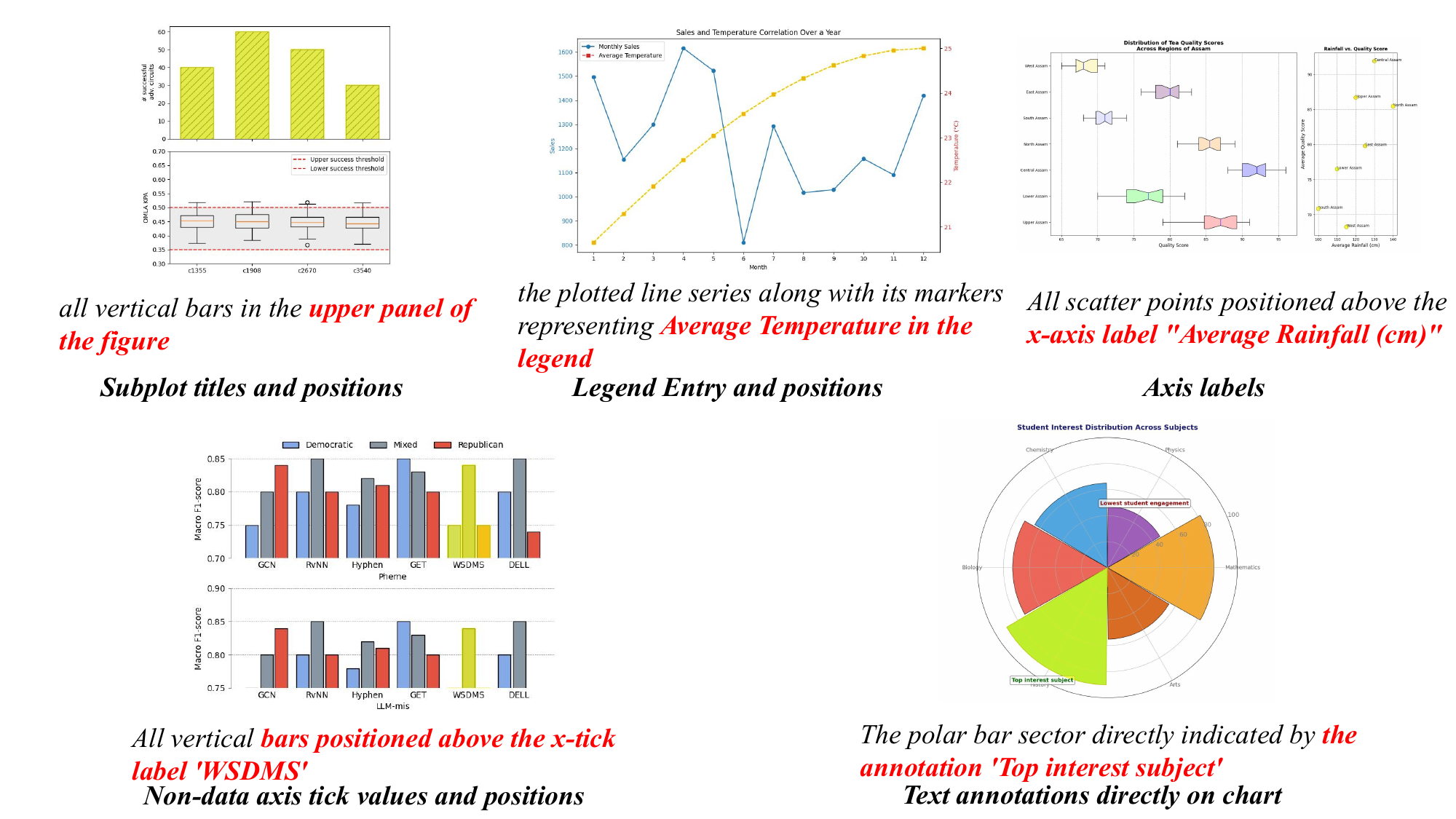}
  \caption{visual referring clue example}
  \label{fig:textual/localization_referring_clue_example}
\end{figure*}

\begin{figure*}[h]
  \centering
  \includegraphics[width=\textwidth]{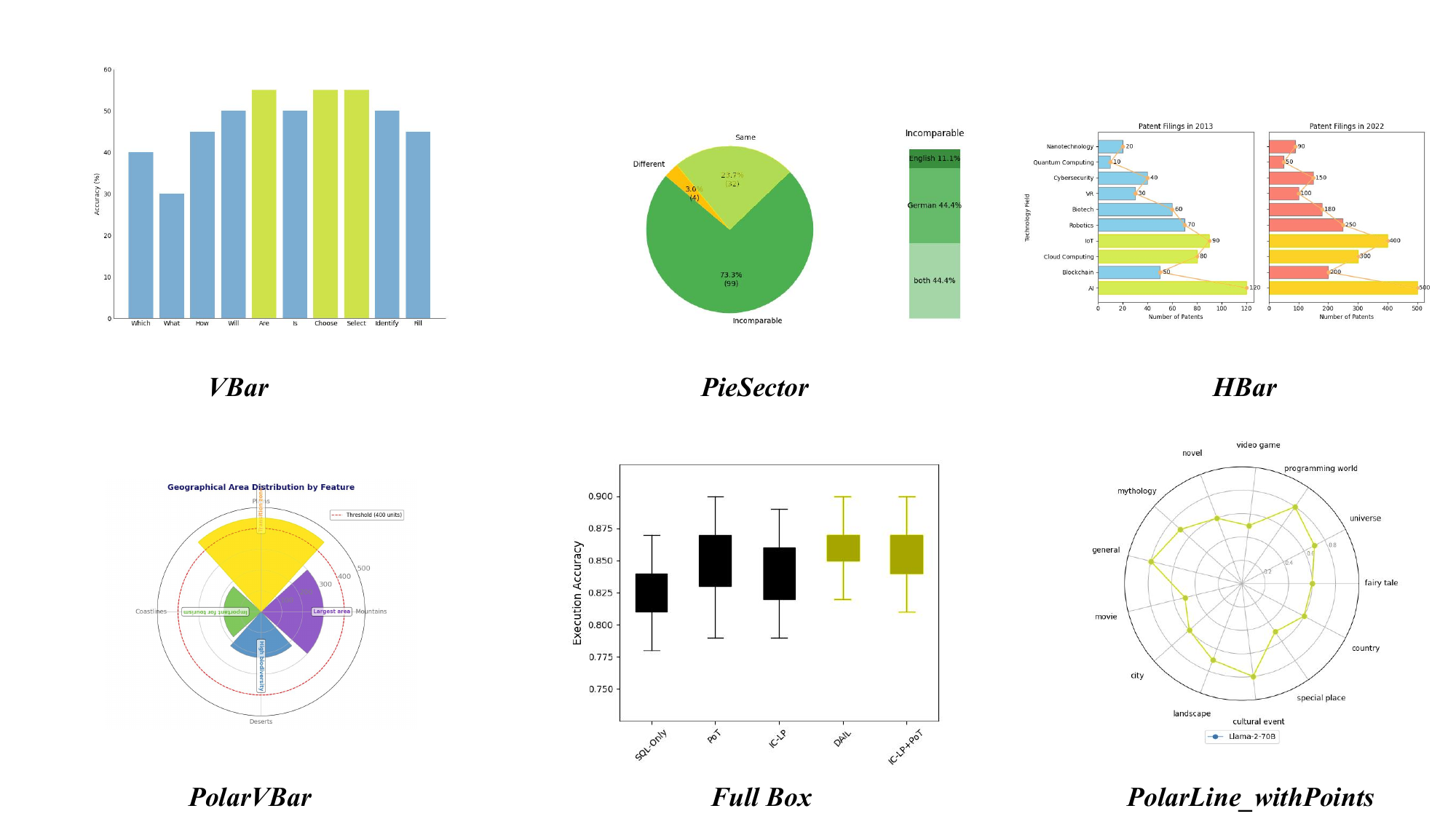}
  \caption{referring target element example}
  \label{fig:per_category_example_1}
\end{figure*}

\begin{figure*}[h]
  \centering
  \includegraphics[width=\textwidth]{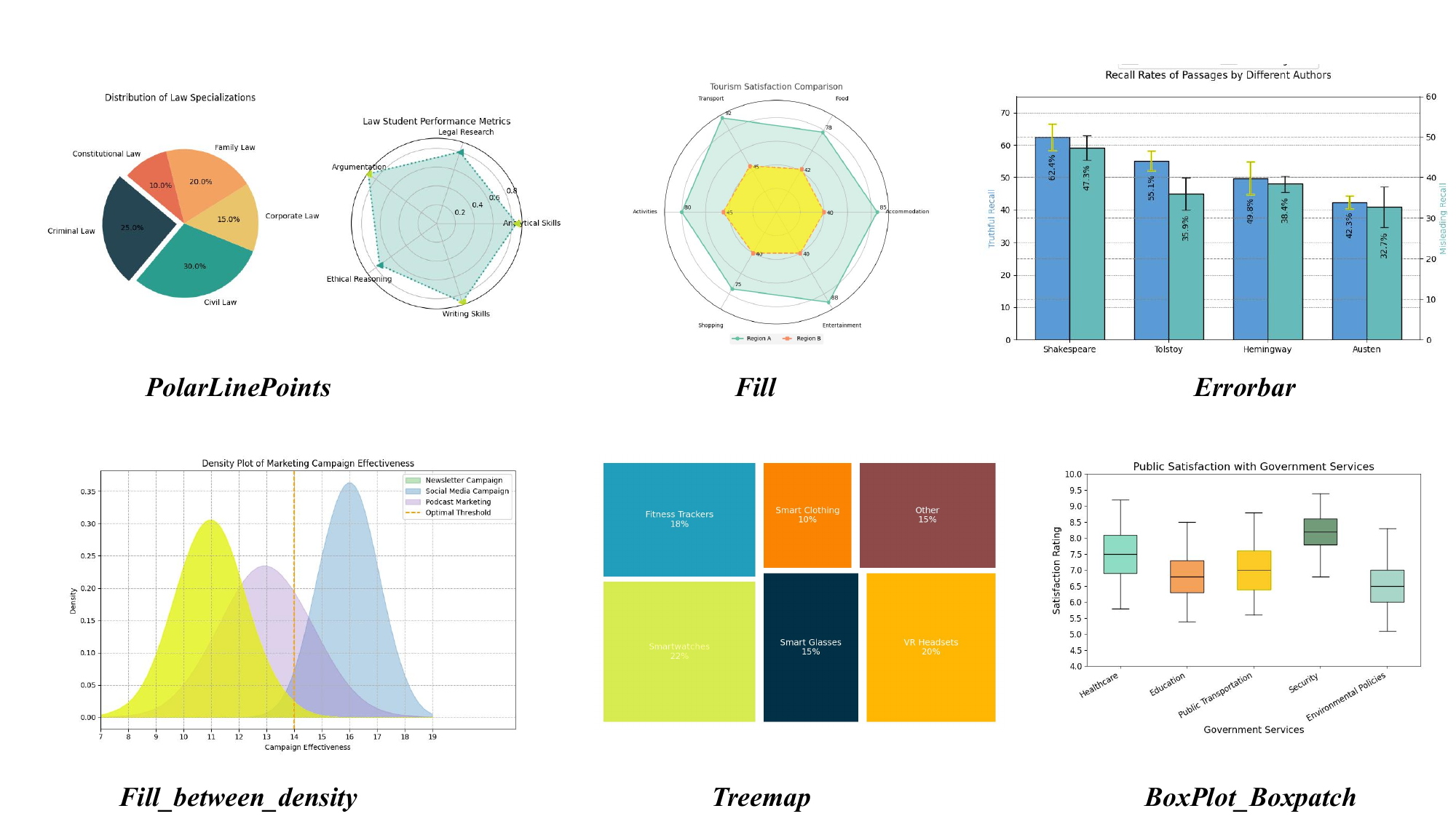}
  \caption{referring target element example}
  \label{fig:per_category_example_2}
\end{figure*}

\begin{figure*}[h]
  \centering
  \includegraphics[width=\textwidth]{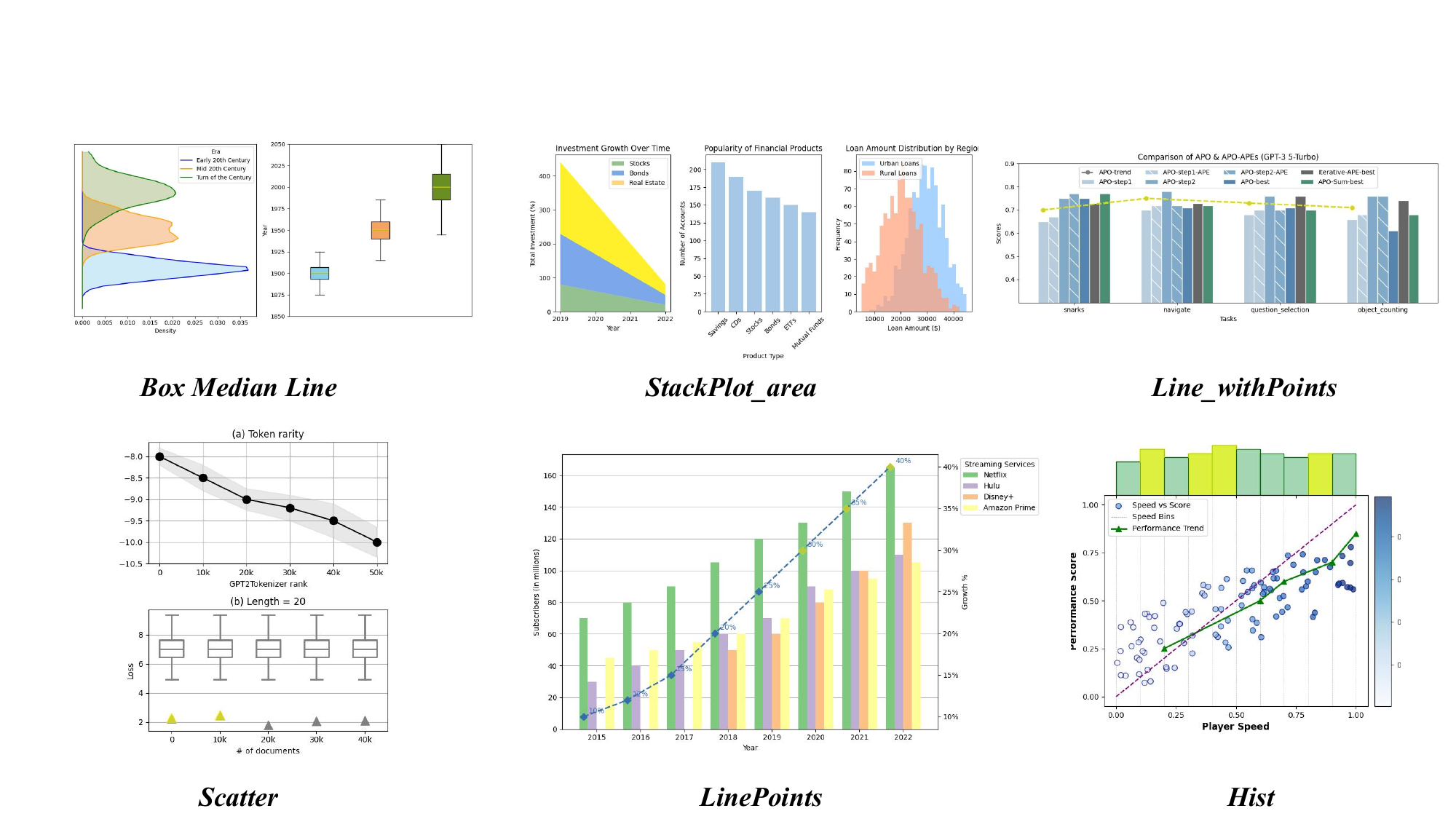}
  \caption{referring target element example}
  \label{fig:per_category_example_3}
\end{figure*}

We show the benchmark examples in Figure~\ref{fig:data_referring_clue_example},
Figure~\ref{fig:visual_referring_clue_example},
Figure~\ref{fig:textual/localization_referring_clue_example} for each referring clues and show examples in Figure~\ref{fig:per_category_example_1}, Figure~\ref{fig:per_category_example_2}, Figure~\ref{fig:per_category_example_3}
for each target element types.

\newpage

\clearpage

\UseRawInputEncoding
\begin{promptbox}[title={Full Prompt for Referring-Expression Generation (Data-related)},label={box:full_prompt_data_related}]
\begin{quote}
\small
Prompt

**You are an expert in data visualization and a professional educator.**

You are asked to generate **visual grounding referring expressions** that describe **some target elements in the chart** based on the given Python code. The descriptions will be used to test a model’s ability to perform **referring expression segmentation (RES/GRES)** with **recognition-style constraints**: the model only needs to recognize targets via **directly observable** cues (comparisons, ordering, local patterns, explicit text anchors, and simple visual attributes), **without any arithmetic/statistics**.

---

The Given Python Code

[python_code]

Target Element Type

[target_element_type]

The target elements are plotted in the code line marked with **#**.

---

Core Task

Generate referring expressions whose referents are chosen **only from the Target Element Type instances plotted by the single code line marked with `#`**. Each referring expression must identify either:

* **(A) a SINGLE instance** of that `#`-plotted Target Element Type, OR
* **(B) a SET of multiple eligible instances** of that `#`-plotted Target Element Type.

---

CRITICAL REQUIREMENTS (Hard)

1) Target Composition (Strict)

* The referent(s) must be **only** the specified **Target Element Type**.
* Do **not** refer to any other graphic element types as targets.

2) Referent Existence (Strict)

* Each referring expression must refer to **at least one** valid target instance in the rendered chart.
* Do not generate “no-target” expressions.
* **Feasibility guard:** Avoid self-contradictory constraints (e.g., mutually exclusive rank/range/pattern conditions) that could plausibly yield an empty set.

3) Referent Subject (Strict)

* Each referring expression must **explicitly begin** with **[referent subject]** as expression start.
* **Format hardening:** The `referring_expression` string must start with **exactly** the characters `[referent subject]` as the **very first characters** (no leading spaces/newlines/punctuation before it).

4) Rendered-Image-Only Constraint (Strict)

* The model answering will see **only the rendered image**, not the code. Therefore:

  * **Never** mention code-level details (variable names, function calls, parameters, hex color codes, random seeds, etc.).
  * If the code intent and the rendered view could differ, describe only what is **visually apparent**.
  * There may be discrepancies between the code specification and the final visual representation (e.g., code might specify sizes, but the rendered pie chart displays percentages). Descriptions must reflect the visual outcome, not the code intent.

5) Random Data Constraint (Strict)

If target data are generated using random functions:

* Use **only** relations implied by explicit random parameters (distribution/bounds/monotonic transforms).
* **Never** cite specific generated numeric values.
* Prefer robust non-empty selection styles (rank-bands, local patterns, tick-anchored ranges) over fragile narrow numeric cutoffs.

6) Recognition-Only Rule (No Calculations) (Strict)

These are **recognition_data** expressions:

* **Do NOT** use arithmetic or statistics: no differences/ratios/rates, no mean/median/std, no “average + …”, no derived thresholds.
* You **may** use:

  * direct comparisons (“higher/lower”, “above/below”) on visible values,
  * rank selection by comparisons (top/bottom/rank band),
  * local adjacency patterns (peaks/troughs/reversals/runs) via pairwise comparisons,
  * cross-series “A above B at the same x” comparisons (**no gap/ratio**).

---

Feature System (Taxonomy)

**Important:** A single referring expression may belong to **multiple categories**.
In your output labels, include **all categories actually used** by the expression.

A) Data Feature Categories (use one or more)

1. **Value-Range Filtering**
   Targets are all elements whose values are **within/above/below** a specified **range/interval**, where boundaries are **directly given constants** or **explicitly referenced in the chart** (ticks/labels/on-mark labels/some reference mark point). **No derived thresholds**.

2. **Rank-Band Set Selection**
   Targets are all elements whose **rank positions** fall in a specified band (top-N, bottom-N, ranks i–j, excluding extremes) within an explicit scope, determined by **ordering/comparisons only** (no arithmetic/statistics).
   **Tie policy:** if ties occur at the boundary, **include all tied elements**.

3. **Local-Structure Patterns**
   Targets are elements defined by **local adjacency comparisons** along a series: local peaks/troughs, reversals, neighbor comparisons, and contiguous increasing/decreasing/plateau runs—**purely by pairwise higher/lower/equal comparisons**, with no computed rates or aggregates.

4. **Cross-Series Relations**
   Targets are defined by **cross-series comparisons** at the same x/category (A above/below B), or per-category winner/loser by comparison, with **no gap/ratio calculations**.
   *If multiple series exist, series identity must be disambiguated using legend text (textual/localization) or visual attributes (visual).*


B) Textual/Localization Feature Categories (for data + textual/localization; use one or more)

1. **Axis labels**
**Definition:** Targets are selected using **explicit axis label text** (e.g., x/y axis titles) as an unambiguous anchor to specify which axis (or which subplot’s axis) the reference applies to.

2. **Non-data axis tick values and positions**
**Definition:** Targets are selected using **Non-data axis tick labels (values) and their positions**  as explicit, non-data anchors—without requiring exact value reading beyond the printed tick text.

3. **Legend entries and their positions**
**Definition:** Targets are selected by **legend entry text** (and optionally its **layout position**, e.g., first/second entry, top/bottom of legend) to map from label → corresponding elements/series in a scope.

4. **Subplot titles, identifiers and positions**
**Definition:** Targets are selected by **subplot-level text identifiers** (e.g., subplot title, facet header label, panel tag like “(a)/(b)”) and/or their **panel positions** to disambiguate which subplot the reference is in.

5. **Text annotations directly on the chart**
**Definition:** Targets are selected using **explicit on-chart text** (callouts, data labels, annotation strings) that is visually attached to marks or regions, serving as a direct textual anchor for grounding.

C) Visual Feature Categories (for data + visual; use one or more)

1.  **Color Attributes**
**Definition:** Targets are elements/series identified by a **discrete color label** (e.g., red/blue/green), not subjective shades.

2.  **Shape Style**
**Definition:** Targets are elements identified by a **fixed, enumerated elements/shape name**, e.g., **circle, square, diamond, cross, plus, x, star, pentagon, hexagon**, and oriented variants such as **triangle-up/down/left/right**.

3.  **Line Style / Stroke Style**
**Definition:** Targets are elements identified by the **stroke/outline pattern** (solid/dashed/dotted/dashdot). This applies to **any visible edge**, including **lines** and **borders/outlines** of bars/areas/markers. 

4.  **Fill Style**
**Definition:** Targets are elements identified by **interior fill appearance**: **filled vs hollow (outline-only)**, and (when present) **hatch/pattern type and direction** (e.g., diagonal/vertical/horizontal/crosshatch).

---

Generation Task (Counts + Mix)

Generate **exactly 20** distinct referring expressions:

* `data_only`: **10** items (Data Feature categories only; no visual/textual/localization)
* `data_plus_visual`: **5** items (must require both data + visual)
* `data_plus_textual_localization`: **5** items (must require both data + textual/localization)

**Interdependence requirement for mixed splits:** removing either feature type should make the reference ambiguous or change the referent set.

**Avoid over-reliance on extremes (soft constraint):**
* Do **not** overuse superlative extreme-rank words like “largest/smallest/highest/lowest/topmost/bottommost”.
* Prefer **multi-object** selections such as **top-N / bottom-N**, broad tick-anchored ranges, local-pattern sets, or per-category comparisons.

**Category diversity requirement (important):**
* Make the **category usage as diverse/different as possible** across the 20 expressions.
* Avoid producing many expressions with the **same category combination** (e.g., identical `data_feature_categories` lists, or identical full combinations including visual/text categories).
* Try to cover **all four** Data Feature categories across the whole set, and vary which Visual/Textual/Localization categories are used in the mixed splits.

**Chart-Contextual Terminology:**

Use terminology that matches the plotted chart and **avoid** generic terms like “data-value” or “element”.

If the chart type has subcomponents (e.g., errorbar, boxplot), specify which component is meant (central point vs. upper/lower bound, Q1/Q3/median, etc.).


**Cross-subplot:** if multiple subplots exist, include **at least 2** expressions that require subplot disambiguation (typically via subplot titles/identifiers/positions or other explicit anchors).

**Distinctness:** all 20 expressions must be meaningfully different (not just paraphrases).

---

Strictly Prohibited

* Any reference to code-level specifications (variable names, parameters, hex color codes, seeds).
* Any arithmetic/statistics/derived computations (difference, ratio, rate, mean/median/std, derived thresholds).
* Subjective visual judgments (“brightest”, “most saturated”, “densest”).

---

OUTPUT

Return **one JSON object** with three arrays:

* `data_only` (10 items)
* `data_plus_visual` (5 items)
* `data_plus_textual_localization` (5 items)

Each item must include:

* `referring_expression` (must start with **[referent subject]**)
* `data_feature_categories` (array, len ≥ 1; values must be exact category names from the Data Feature taxonomy above)

Only when applicable:

* `visual_feature_categories` (array, len ≥ 1; values from the Visual taxonomy above)
* `textual_localization_categories` (array, len ≥ 1; values from the Textual/Localization taxonomy above)

Category arrays may contain **multiple labels**; include **all** categories actually used.

Minimal template:

```json
{
  "data_only": [
    {"referring_expression": "...", "data_feature_categories": ["..."]}
  ],
  "data_plus_visual": [
    {"referring_expression": "...", "data_feature_categories": ["..."], "visual_feature_categories": ["..."]}
  ],
  "data_plus_textual_localization": [
    {"referring_expression": "...", "data_feature_categories": ["..."], "textual_localization_categories": ["..."]}
  ]
}
```

Task
Please carefully analyze the requirements of the prompt step by step and then generate the output json.
\end{quote}
\end{promptbox}

\clearpage
\UseRawInputEncoding
\begin{promptbox}[title={Full Prompt for Referring-Expression Generation (Non-Data-related)},label={box:full_prompt_non_data_related}]
\begin{quote}
\small
# Prompt

**You are an expert in data visualization and a professional educator.**

You are asked to generate **visual grounding referring expressions** that describe **some target elements in the chart** based on the given Python code. The descriptions will be used to test a model’s ability to perform **referring expression segmentation (RES/GRES)** with **recognition-style constraints**: the model only needs to recognize targets via **directly observable** cues (explicit text anchors, objective spatial localization, and simple visual attributes), **without any arithmetic/statistics**.

---

## The Given Python Code

[python_code]

## Target Element Type

[target_element_type]

The target elements are plotted in the code line marked with **#**.

---

## Core Task

Generate referring expressions whose referents are chosen **only from the Target Element Type instances plotted by the single code line marked with `#`**. Each referring expression must identify either:

* **(A) a SINGLE instance** of that `#`-plotted Target Element Type, OR
* **(B) a SET of multiple eligible instances** of that `#`-plotted Target Element Type.

---

## CRITICAL REQUIREMENTS (Hard)

### 1) Target Composition (Strict)

* The referent(s) must be **only** the specified **Target Element Type**.
* Do **not** refer to any other graphic element types as targets.

### 2) Referent Existence (Strict)

* Each referring expression must refer to **at least one** valid target instance in the rendered chart.
* Do not generate “no-target” expressions.
* **Grounding feasibility guard:** Use the Python code only to understand what target instances actually exist (how many, where they appear, which subplots they are in). Do **not** invent targets that are not plotted by the single `#` line. Avoid self-contradictory constraints that could plausibly yield an empty set.

### 3) Referent Subject (Strict) 

* Each referring expression must **explicitly begin** with **[referent subject]** as expression start.
* **Format hardening:** The `referring_expression` string must start with **exactly** the characters `[referent subject]` as the **very first characters** (no leading spaces/newlines/punctuation).

### 4) Rendered-Image-Only Constraint (Strict)

* The model answering will see **only the rendered image**, not the code. Therefore:

  * **Never** mention code-level details (variable names, function calls, parameters, hex color codes, random seeds, etc.).
  * If the code intent and the rendered view could differ, describe only what is **visually apparent**.
  * There may be discrepancies between the code specification and the final visual representation (e.g., code might specify sizes, but the rendered pie chart displays percentages). Descriptions must reflect the visual outcome, not the code intent.

### 5) Recognition-Only Rule (No Calculations) (Strict)

These are **recognition_data** expressions:

* **Do NOT** use arithmetic or statistics: no differences/ratios/rates, no mean/median/std, no “average + …”, no derived thresholds.

### 6) Random Data Constraint (Strict)

If target data are generated using random functions:

* Use **only** relations implied by explicit random parameters (distribution/bounds/monotonic transforms).
* **Never** cite specific generated numeric values.

---

## Feature System (Taxonomy)

**Important:** A single referring expression may belong to **multiple categories** within its feature type(s).
In your output labels, include **all categories actually used** by the expression.

### A) Textual/Localization Feature Categories (use one or more when text/loc is allowed)

1. **Axis labels**
**Definition:** Targets are selected using **explicit axis label text** (e.g., x/y axis titles) as an unambiguous anchor to specify which axis (or which subplot’s axis) the reference applies to.

2. **Non-data axis tick values and positions**
**Definition:** Targets are selected using **Non-data axis tick labels (values) and their positions**  as explicit, non-data anchors—without requiring exact value reading beyond the printed tick text.

3. **Legend entries and their positions**
**Definition:** Targets are selected by **legend entry text** (and optionally its **layout position**, e.g., first/second entry, top/bottom of legend) to map from label → corresponding elements/series in a scope.

4. **Subplot titles, identifiers and positions**
**Definition:** Targets are selected by **subplot-level text identifiers** (e.g., subplot title, facet header label, panel tag like “(a)/(b)”) and/or their **panel positions** to disambiguate which subplot the reference is in.

5. **Text annotations directly on the chart**
**Definition:** Targets are selected using **explicit on-chart text** (callouts, data labels, annotation strings) that is visually attached to marks or regions, serving as a direct textual anchor for grounding.


### B) Visual Feature Categories (use one or more when visual is allowed)

1.  **Color Attributes**
**Definition:** Targets are elements/series identified by a **discrete color label** (e.g., red/blue/green), not subjective shades.

2.  **Shape Style**
**Definition:** Targets are elements identified by a **fixed, enumerated elements/shape name**, e.g., **circle, square, diamond, cross, plus, x, star, pentagon, hexagon**, and oriented variants such as **triangle-up/down/left/right**.

3.  **Line Style / Stroke Style**
**Definition:** Targets are elements identified by the **stroke/outline pattern** (solid/dashed/dotted/dashdot). This applies to **any visible edge**, including **lines** and **borders/outlines** of bars/areas/markers. 

4.  **Fill Style**
**Definition:** Targets are elements identified by **interior fill appearance**: **filled vs hollow (outline-only)**, and (when present) **hatch/pattern type and direction** (e.g., diagonal/vertical/horizontal/crosshatch).  

---

## Generation Task (Counts + Mix)

Generate **exactly 15** distinct referring expressions:

* **textual_localization_only**: **5** items
  * Use **ONLY** Textual/Localization features (no visual attributes).
  * Do not use color/shape/line style/fill style.
  * Expressions must remain resolvable from explicit text anchors or anchored localization only.

* **visual_only**: **5** items
  * Use **ONLY** Visual features (no textual/localization cues).
  * Do not mention legend text, axis labels, tick labels, subplot titles, annotations, or anchored location references.

* **textual_localization_plus_visual**: **5** items
  * Must require **BOTH** Textual/Localization and Visual features to identify the referent.
  * **Interdependence:** removing either the text/loc part or the visual part should make the reference ambiguous or change the referent set.

✅ **Avoid over-reliance on extremes (soft constraint):**
* Do **not** overuse superlative/extreme wording (e.g., “leftmost/rightmost/topmost/bottommost”, “the only one”, “the single most …”) as the primary disambiguator.
* Prefer **multi-target** selections and/or combined cues.

✅ **Category diversity requirement (important):**
* Make category usage **as diverse/different as possible** across the 20 expressions.
* Avoid producing many expressions with the **same category combination** (e.g., identical `textual_localization_categories` lists, identical `visual_feature_categories` lists, or identical full combinations in the mixed split).
* Across the whole set, try to cover **all** available Textual/Localization categories and **all** available Visual categories whenever the chart makes them applicable.

**Chart-Contextual Terminology:**
Use terminology that matches the plotted chart and **avoid** generic terms like “data-value” or “element”.
If the chart type has subcomponents (e.g., errorbar, boxplot), specify which component is meant (central point vs. upper/lower bound, Q1/Q3/median, etc.).

**Cross-subplot:** if multiple subplots exist, include **at least 2** expressions (any split where text/loc is allowed) that explicitly disambiguate the subplot (by title/identifier/position) before identifying targets.

**Distinctness:** all 20 expressions must be meaningfully different (not just paraphrases).

---

## Strictly Prohibited

* Any reference to code-level specifications (variable names, parameters, hex color codes, seeds).
* Any arithmetic/statistics/derived computations.
* Subjective visual judgments (“brightest”, “most saturated”, “densest”).

---

## OUTPUT

Return **one JSON object** with three arrays:

* `textual_localization_only` (5 items)
* `visual_only` (5 items)
* `textual_localization_plus_visual` (5 items)

Each item must include:

* `referring_expression` (must start with **[referent subject]**)

And the category labels:

* If the item uses textual/localization features, include `textual_localization_categories` (array, len ≥ 1; exact names from the taxonomy above).
* If the item uses visual features, include `visual_feature_categories` (array, len ≥ 1; exact names from the taxonomy above).

Category arrays may contain **multiple labels**; include **all categories actually used**.

Minimal template:

```json
{
  "textual_localization_only": [
    {
      "referring_expression": "...",
      "textual_localization_categories": ["..."]
    }
  ],
  "visual_only": [
    {
      "referring_expression": "...",
      "visual_feature_categories": ["..."]
    }
  ],
  "textual_localization_plus_visual": [
    {
      "referring_expression": "...",
      "textual_localization_categories": ["..."],
      "visual_feature_categories": ["..."]
    }
  ]
}
```

# Task
Please carefully analyze the requirements of the prompt step by step and then generate the output json.
\end{quote}
\end{promptbox}

\clearpage
\UseRawInputEncoding
\begin{promptbox}[title={Referring expression target generation prompt},label={box:full_prompt_ref_target_gen}]
\begin{quote}
\small
# Prompt:

## Core task

Given:
1) a detailed target element type description,
2) a referring expression (natural language) that refers to one or multiple elements of that target type in the final plot,
3) Python code that generates the visualization,

Return which visual element(s) are referred to by the referring expression, **restricted to the target elements created at the code lines marked with `#`**.

You MUST reason about the **final visual appearance** after the entire code finishes executing (including axis scaling, normalization, transforms, limits, z-order, styling, and any later modifications to previously created artists), not raw data values.

## Inputs

**Target Element Type (detailed):**
[target_element_type]

**Referring Expression:**
[ref_sentence]

**Given Code (markers `#1`, `#2`, ... appear only on lines that create target elements):**
```python
[python_code]
```

## Additional Rules

### 1) Marked line execution & invocation_count

* Each marked line (e.g., `#1`) may run multiple times (e.g., inside loops).
* For each marked line, maintain an independent counter starting at 0:

  * The 1st time `#1` runs → `invocation_count = 0`
  * The 2nd time `#1` runs → `invocation_count = 1`
  * etc.
* Count separately per marker (`#1`, `#2`, ...).

### 2) element_indices (indexing within one execution)

For a specific execution (a specific `line` + `invocation_count`):

* If one execution of the marked line creates **k > 1** target elements, assign indices **0..k-1 strictly by the order of the input data passed in that execution of the plotting call** (i.e., the sequence order in the code’s iterable).
* If that execution creates exactly **one** target element of the target element type and it matches the referring expression, output `"element_indices": null`.
* If that execution creates multiple elements of the target element type and only some match, output the list of their indices.
* If none match, do not output an entry for that execution.

## Output format

You should return **ALL** the referring target elements result in the following JSON format:

```json
{
    "results": [
        {
            "line": "#1",
            "invocation_count": 2,
            "element_indices": null
        },
        {
            "line": "#1",
            "invocation_count": 5,
            "element_indices": [0, 1, 2]
        },
        {
            "line": "#2",
            "invocation_count": null,
            "element_indices": null
        }
    ]
}
```

**Field Explanations:**

- `line` (string, required):
    - The code line marker (e.g., `"#1"`, `"#2"`) that create at least one reference to a target visual element.
    - This field should include **only** lines that actually create or refer to at least one of the target elements.
- `invocation_count`(integer, starting from 0):
    - **Which time** the marked line was executed.
    - Count separately for each marked line (`#1`, `#2`, etc.).
- `element_indices`(list of integers or `null`):
    - **If multiple** target elements were created in that execution and satisfy the expression: list their **0-based indices**.
    - **If only one** target element was created and satisfies the expression: set to `null`.
    - **If no** elements satisfy the expression: **do not include** that execution in the results.


# Task:

Please carefully read the prompt and reason step by step carefully and then providing your final output json dicts.
\end{quote}
\end{promptbox}

\end{document}